\documentclass[twocolumn]{svjour3}          
\smartqed  
\usepackage{lineno,hyperref}
\usepackage{amssymb}
\usepackage{epsfig}
\usepackage{graphicx,subfigure}
\usepackage{graphbox}
\usepackage{amsmath}
\usepackage[ruled, linesnumbered]{algorithm2e}
\usepackage{bm}
\usepackage{rotating}
\usepackage{cases}
\usepackage{nccbbb}
\usepackage{epsfig}
\usepackage{booktabs}
\usepackage[table,xcdraw]{xcolor}
\usepackage{multicol}
\usepackage{tabularx}
\usepackage[misc]{ifsym}
\usepackage{changes}
\usepackage[font=footnotesize,labelfont=bf]{caption}
\usepackage{lscape}
\usepackage{longtable}
\usepackage{comment}
\usepackage{ulem}

\usepackage{array}
\usepackage{textcomp}
\usepackage{stfloats}
\usepackage{url}
\usepackage{verbatim}
\usepackage{graphicx}
\usepackage{cite}
\usepackage{color}
\usepackage{booktabs}
\usepackage{mathrsfs}
\usepackage{footnote}
\usepackage{enumerate}
\usepackage{float}
\usepackage{upgreek}
\usepackage{multirow}
\usepackage{makecell}
\usepackage{pifont}
\usepackage{algorithmic}

\graphicspath{{./figure/}}

\begin{document}

\title{Win-Win Fusion: Multi-Modal Image Fusion with Visual-Semantic Collaboration}

\author{Hao Zhang, Xuhui Zuo, and Jiayi Ma}


\institute{H. Zhang\and
	        X. Zuo \and
            J. Ma (\Letter)\at
            Electronic Information School, Wuhan University, Wuhan 430072, China
            \\
            \email{zhpersonalbox@gmail.com, \\
            	 xuhuizuo2001@whu.edu.cn,
            	 jyma2010@gmail.com}
}

\date{Received: date / Accepted: date}

\maketitle

\begin{abstract}
Existing multi-modal image fusion (MMIF) methods generally face an optimization gap between the visual and semantic domains, limiting the enhancement of perception capabilities. To overcome this long-standing barrier, this work proposes W2F, a new visual-semantic mutual reinforcing paradigm for MMIF. First, we reinforce the foundation for visual-semantic collaboration by amplifying feature complementarity through two targeted designs. Specifically, the segmentation-guided fusion loss introduces object-aware structural guidance into visual fusion, while the embedded information restoration prior strengthens the ability of visual fusion to provide reliable perceptual evidence for the semantic branch under degraded conditions. Then, we introduce a strong multi-modal information interaction mechanism composed of interactive gated mixed attention (IGM-Att) and progressive coupling attention (PC-Att). These modules ensure comprehensive scene representation from both local and global perspectives, providing informational support for visual-semantic collaboration. Finally, we construct a collaborative mechanism to couple visual fusion and semantic fusion. It employs a feature search module to extract complementary key features from both tasks, enriching the representational potential of visual and semantic information. Meanwhile, a gradient correction strategy ensures collaboration from the optimization perspective, enhancing performance for both visual and semantic fusion through consistency learning. Extensive experiments demonstrate that W2F surpasses state-of-the-art methods in both visual appeal and decision accuracy. The code is publicly available at \url{https://github.com/HaoZhang1018/W2F}.

\end{abstract}

\section{Introduction}\label{Sec:intro}
Due to limitations inherent in imaging principles, a single sensor can only capture information related to a specific aspect of the imaging scene~\cite{zhang2023visible}. In other words, it fails to support the need for comprehensive scene representation, which is particularly pronounced under harsh conditions. As a result, this not only hinders visual perception for human users but also makes machine semantic understanding and decision-making more challenging. Against this backdrop, multi-modal image fusion (MMIF) emerges, enhancing perceptual capabilities by combining image data captured by various sensors, both visually and semantically~\cite{zhang2021image,xu2023murf}. To date, this technology has been widely applied in various fields, \textit{e.g.}, security monitoring~\cite{drajic2007adaptive}, and night-assisted driving~\cite{gong2023sifdrivenet}, demonstrating significant value.

According to the technical goal of MMIF, perception enhancement can be understood from both visual and semantic aspects~\cite{liu2024infrared, wang2026advances}. This naturally derives two specific fusion tasks. i) \textit{\textbf{Visual fusion}}: it is aimed at generating a single fused image with rich texture details and significant contrast, with the intention of improving visual perception~\cite{li2023lrrnet,zhou2024probing}. ii) \textit{\textbf{Semantic fusion}}: it focuses on enhancing machine understanding, strengthening decision-making, and improving the description of the imaging scene, \textit{e.g.}, semantic segmentation~\cite{lv2024context,zhao2024open}, object detection~\cite{tang2024divide,fu2023lraf}. These two fusion tasks form a complete perception enhancement system, spanning from visual to semantic levels.

\begin{figure*}[t]
	\centering
	\includegraphics[width=1\linewidth]{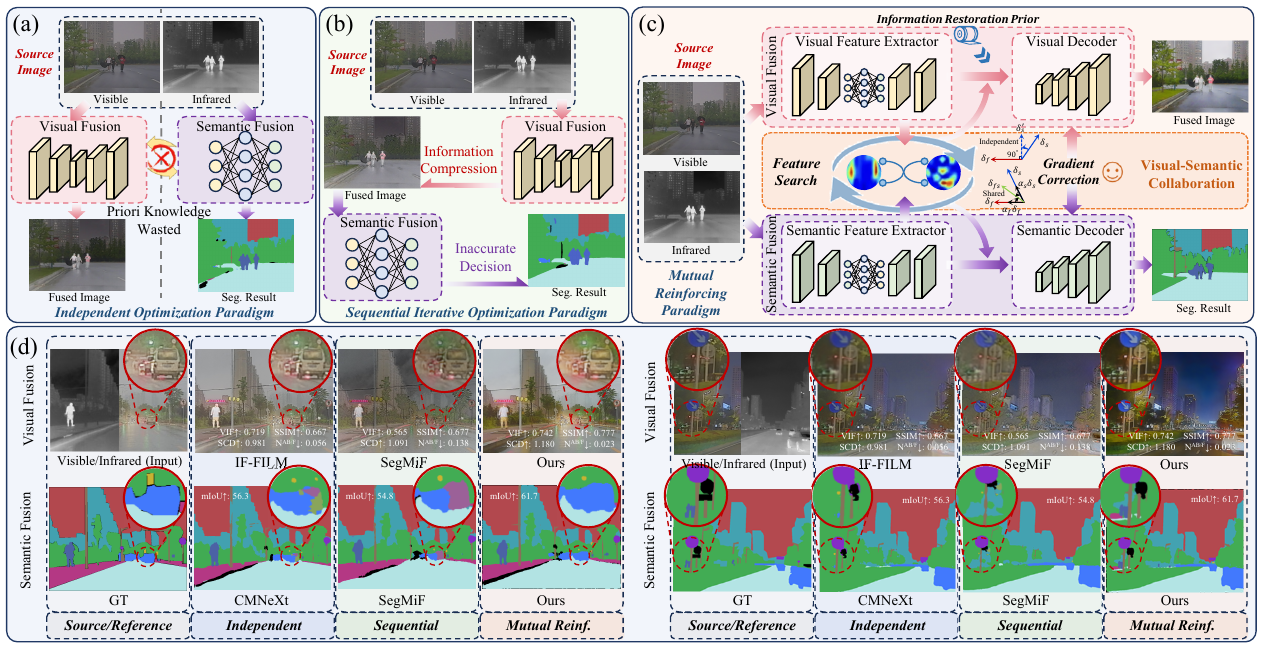}
	\vspace{-0.2in}
	\caption{Illustration of three optimization paradigms for the interaction between visual and semantic fusion. (a) Independent optimization paradigm; (b) sequential iterative optimization paradigm; (c) our proposed mutual reinforcing paradigm; (d) comparison of results from different paradigms, where IF-FILM~\cite{zhao2024image} and CMNeXt~\cite{zhang2023delivering} represent the independent optimization paradigm, and SegMiF~\cite{liu2023multi} represents the sequential iterative optimization paradigm. The source images are a paired visible and infrared image sample from the FMB dataset~\cite{liu2023multi}.}\label{fig1}
\end{figure*}

Several MMIF methods have emerged in recent years to achieve visual fusion and semantic fusion, which can be categorized into two paradigms. i) \textit{\textbf{Independent optimization paradigm}}: it generally treats visual fusion and semantic fusion as two separate processes~\cite{li2024deep,yi2024text,lv2024context}, as shown in Fig.~\ref{fig1} (a). Thus, methods of this type tend to carry out methodology design according to the specific characteristics of these two tasks, respectively. This makes the potential prior knowledge provided by one task unavailable to the other, resulting in inevitable performance bottlenecks for both. ii) \textit{\textbf{Sequential iterative optimization paradigm}}: it regards semantic fusion as the next step of visual fusion, and enforces a cascade dependency between the two tasks~\cite{tang2022image,liu2023multi}, as shown in Fig.~\ref{fig1} (b). However, visual fusion is essentially a process of \textbf{information compression}, which may discard specific cues useful for semantic understanding. Consequently, the semantic fusion model receives incomplete information for decision-making, and inaccurate semantic decisions may in turn constrain the optimization of visual fusion.

Clearly, these two existing MMIF paradigms fail to fully exploit the intrinsic complementarity between visual fusion and semantic fusion. Then, where does such complementarity originate, and what benefits can it bring? To answer this question, we analyze it from the perspective of feature representation. Specifically, scene understanding can be viewed as a progressive abstraction from visual perceptual evidence to object-level semantic representations. The distinct optimization objectives of visual fusion and semantic fusion drive them to focus on different yet complementary representations. Visual fusion mainly emphasizes perceptual evidence such as textures, contrast, boundaries, and structural details, whereas semantic fusion focuses more on object regions, foreground-background relationships, and category-discriminative representations for accurate scene understanding. Therefore, visual fusion can provide visibility-restoration-oriented visual perceptual evidence to support semantic fusion in further constructing reliable object-level representations under degraded conditions, while semantic fusion can provide object-aware structural priors to guide visual fusion toward semantically important objects. Such reciprocal interactions give rise to an intrinsic bidirectional complementarity between visual and semantic representations.

A natural question therefore arises: can we simultaneously achieve superior visual fusion quality and semantic fusion performance by fully exploiting such complementarity? This motivates the exploration of a collaborative framework in which visual fusion and semantic fusion actively reinforce each other, rather than being treated independently or sequentially. By enabling bidirectional information exchange and mutual guidance between the two tasks, such a framework has the potential to bridge the optimization gap between the visual and semantic domains and further improve the overall perception capability.

However, achieving this goal is still highly challenging. Firstly, although the complementarity between visual fusion and semantic fusion exists inherently, it is still too weak to effectively support deep visual-semantic collaboration. Secondly, there is a lack of powerful multi-modal information interaction mechanisms to complement scene features at both the visual and semantic levels. This leaves visual-semantic collaboration lacking fundamental informational support, let alone the possibility of mutual reinforcement. Therefore,  \textit{amplifying the visual-semantic complementarity and constructing a high-quality multi-modal information interaction mechanism} are fundamental to achieving effective mutual reinforcement between visual fusion and semantic fusion.

We attempt to solve the above two challenges within a new mutual reinforcing paradigm for MMIF, termed W2F, as shown in Fig.~\ref{fig1} (c). In this work, \textit{\textbf{semantic segmentation is selected as the representative task for semantic fusion to concretize our methodology}}. Firstly, we amplify the complementarity between visual fusion and semantic segmentation through two targeted designs. Specifically, considering that segmentation emphasizes semantic characteristics, such as object regions and structural boundaries, we design a segmentation-guided fusion loss to introduce object-aware structural guidance into visual fusion, thereby improving object visibility and distinguishability. Meanwhile, we embed an information restoration prior into the optimization objective of visual fusion, strengthening its capability to extract visibility-restoration-oriented visual perceptual evidence from degraded observations, thereby enhancing its ability to support the semantic branch in constructing reliable object-level representations under degraded conditions. Secondly, we design two attention modules, CNN-based interactive gated mixed attention (IGM-Att) and Transformer-based progressive coupling attention (PC-Att), to implement high-quality interaction of multi-modal information. Among them, IGM-Att integrates a trainable gating mechanism to identify misjudged positional features and correct mismatched high or low scores~\cite{zhang2024mrfs}, thereby achieving cross-modal visual information completion from a local perspective. On this basis, PC-Att achieves sufficient semantic aggregation from a global perspective through the progressive coupling of single-modal self-reinforcement and cross-modal mutual complementarity. Together, they provide strong informational support for visual-semantic collaboration. Furthermore, we develop a collaborative mechanism to couple visual fusion and segmentation. Concretely, we develop a feature search module to extract complementary key features from these two tasks, bridging the gap between visual and semantic information and enriching their representational potential. Such an injection of prior knowledge effectively optimizes the information domain on which the visual fusion and segmentation models rely. Meanwhile, a gradient correction strategy is designed to ensure collaboration between the two tasks from the optimization perspective. By ensuring learning consistency, it enables visual fusion and segmentation to be mutually reinforcing rather than interfering.

Fig.~\ref{fig1} (d) provides an intuitive comparison among different MMIF paradigms. Existing paradigms either generate fused images with insufficient structural details or produce incomplete semantic predictions due to inadequate prior knowledge utilization. In contrast, W2F achieves clearer structures and more accurate semantic predictions, demonstrating the effectiveness of visual-semantic mutual reinforcement.

To conclude, the contributions of this work include the following five aspects:
\begin{itemize}
	\item[$\bullet$] We propose a mutual reinforcement paradigm between visual fusion and segmentation via visual-semantic collaboration, reducing the optimization gap between the visual and semantic domains in the perception enhancement system. Beyond conventional sequential connections, W2F enables bidirectional interaction between task-specific visual and semantic representations to exploit their complementarity and mutually enhance the two tasks.
	\item[$\bullet$] We amplify the complementarity between visual fusion and semantic segmentation through a segmentation-guided fusion loss and an embedded information restoration prior. In particular, the former provides object-aware structural guidance to enhance object visibility and distinguishability in fused images, while the latter strengthens the ability of visual fusion to provide visibility-restoration-oriented visual perceptual evidence for semantic segmentation under degraded conditions.
	\item[$\bullet$] An interactive gated mixed attention module and a progressive coupling attention module are developed to form a powerful multi-modal information interaction mechanism. They address mismatched feature validity scores and strengthen the sufficiency of feature aggregation, providing high-quality information assurance for visual-semantic collaboration.
	\item[$\bullet$] We construct a collaborative mechanism to couple visual fusion and segmentation, which bridges the gap between visual and semantic information through a feature search module and ensures the consistency of learning between these two tasks using a gradient correction strategy. Therefore, visual fusion and semantic segmentation reinforce each other rather than conflict.
	\item[$\bullet$] We build new RGB-T datasets containing paired clean images based on existing datasets, which can support the community's future development of robust MMIF algorithms.
\end{itemize}

A preliminary version of this paper is MRFS~\cite{zhang2024mrfs}. The new contributions mainly lie in three aspects. First, W2F further analyzes the intrinsic complementarity between visual fusion and semantic segmentation from the perspective of feature representation, and amplifies such complementarity through two targeted designs. Specifically, W2F designs a segmentation-guided fusion loss that is absent in MRFS, enabling segmentation to enhance the visibility and distinguishability of objects in the fused image. Simultaneously, W2F introduces a more powerful information restoration prior into visual fusion to strengthen its capability of providing reliable perceptual evidence for semantic reasoning, which significantly expands the handled degradation types. In particular, MRFS only considers low light in the visible modality and low contrast in the infrared modality. In contrast, our proposed W2F further considers a broader range of composite degradation, including low light, noise, color distortion, and haze in the visible modality, as well as noise and low contrast in the infrared modality. Therefore, W2F demonstrates stronger robustness in the visual fusion task and has a stronger ability to facilitate semantic perception under degraded conditions. Second, W2F fine-tunes the PC-Att module by removing the feedback loop, transforming it into a progressive coupling mode that integrates single-modal self-enhancement and cross-modal complementarity. This simplification reduces the learning difficulty of multi-modal information fusion, enhancing the representational capability. In addition, W2F also improves the collaborative mechanism between visual fusion and semantic segmentation. Concretely, MRFS achieves a hard coupling of these two tasks through feature sharing. By comparison, W2F uses a feature search module and a gradient correction strategy to form a more intelligent collaborative mechanism. It bridges the gap between visual and semantic information while enabling mutual reinforcement between visual fusion and semantic segmentation at the optimization level. Thus, W2F achieves significant performance improvement over MRFS in both tasks.

\section{Related Work}
\subsection{Independent Optimization-based MMIF}
MMIF methods following the independent optimization paradigm advocate treating visual and semantic fusion as two unrelated tasks. Therefore, these methods are typically designed within their own domain of knowledge to achieve their respective goals. In the visual fusion task, there is no ground truth. Thus, existing methods generally treat it as a typical unsupervised optimization problem, with a focus on designing optimization guidance that prioritizes visual quality. For example, U2Fusion~\cite{xu2022u2fusion} and SDNet~\cite{zhang2021sdnet} define visual fusion as a proportional maintenance problem of pixel intensity and gradient. Its motivation is to use pixel intensity to represent contrast and gradients to represent texture, thereby enabling the employment of preferential weighting to selectively integrate these two visual advantages. Differently, FusionGAN~\cite{ma2019fusiongan} and TCGAN~\cite{zhang2023transformer} model visual fusion as an adversarial game, driving the fused image to incorporate visual characteristics from both modalities through unsupervised estimation of multi-source distributions. In the semantic fusion task, decision labels are available to provide explicit optimization guidance. In this context, current methods generally focus on designing neural networks with stronger nonlinear fitting capabilities that enable better interaction between modalities. Typically, CAINet~\cite{lv2024context} proposes a context-aware complementary reasoning module. It establishes the complementary relationship between multi-modal features with the long-term context in both spatial and channel dimensions, exhibiting good capabilities in comprehensive scene understanding. CMX~\cite{zhang2023cmx} employs a cross-modal feature rectification module and a two-stage feature fusion module to gradually achieve the completion and aggregation of scene information, significantly enhancing the global reasoning ability of the segmentation model. However, whether in visual or semantic fusion tasks, being limited to their respective domains of knowledge will inevitably lead to solution limitations. This means that the MMIF method using the independent optimization paradigm wastes the beneficial solution prior that visual and semantic fusion could provide to each other.

\subsection{Sequential Iterative Optimization-based MMIF}
Currently, several methods consider visual and semantic fusion to be sequentially dependent. In other words, they strictly adhere to the notion that visual fusion is a prerequisite step for the semantic fusion task. In this context, a commonly followed sequential iterative optimization paradigm emerges, which advocates incorporating the semantic decision into the optimization guidance of visual fusion. SeAFusion~\cite{tang2022image} and SuperFusion~\cite{tang2022superfusion} are pioneers of this paradigm, which concatenate visual fusion and segmentation models in series. By weighting the appearance fusion loss and segmentation loss and iteratively updating them, both the visual quality and semantic properties of the fused image are effectively improved. However, SeAFusion and SuperFusion use fixed weighting coefficients for these two loss terms, which makes them prone to causing optimization imbalance between visual fusion and segmentation. To mitigate this challenge, SegMiF~\cite{liu2023multi} innovatively introduces dynamic weighting factors based on the descent speed of the loss terms. Intuitively, it adaptively assigns greater weights to tasks with slower learning speeds to help them catch up with the other, thereby allowing semantic segmentation to effectively guide visual fusion and progress together. Similar work also includes TarDAL~\cite{liu2022target} and Detfusion~\cite{sun2022detfusion}. Overall, due to the preliminary consideration and utilization of the relationship between visual and semantic fusion, the above methods achieve improved perception enhancement capabilities. Nevertheless, an unavoidable limitation is that using the result of visual fusion as the input for semantic fusion reduces the scene information available for semantic understanding, as visual fusion is a compression process. Subsequently, the negatively impacted semantic fusion model will, in turn, constrain the optimization of the visual fusion model, leading the entire perceptual enhancement system into an optimization trap. Therefore, exploring a more intelligent interaction mechanism between visual and semantic fusion is highly desirable.

\subsection{Visual-Semantic Collaboration in Other Tasks}
Visual-semantic collaboration has been preliminarily explored in several vision tasks beyond MMIF. TemCoCo~\cite{gong2025temcoco} introduces visual-semantic collaboration into multi-modal video fusion, where semantic representations distilled from foundation models are used to enhance the quality and temporal consistency of fused videos. VSPCN~\cite{jiang2025visual} performs visual and semantic prompt collaboration for generalized zero-shot learning, using semantic attributes as class-level prompts to align visual representations for recognition. MATRN~\cite{na2022multi} develops interactive enhancement between visual and semantic features for text recognition, where linguistic or character-level semantic knowledge is used to compensate for ambiguous visual clues. Although these methods demonstrate the potential of visual-semantic interaction, such collaboration is mainly established within a single-task pipeline, where visual and semantic representations ultimately serve the same optimization objective. In contrast, W2F focuses on MMIF and explicitly couples visual fusion with semantic segmentation within a unified perception enhancement framework. By amplifying visual-semantic complementarity, enabling feature-level interaction, and coordinating the optimization between the two MMIF tasks, W2F achieves bidirectional mutual reinforcement while simultaneously improving fused image quality and pixel-level semantic accuracy.

\begin{figure*}[t]
	\centering
	\includegraphics[width=1\linewidth]{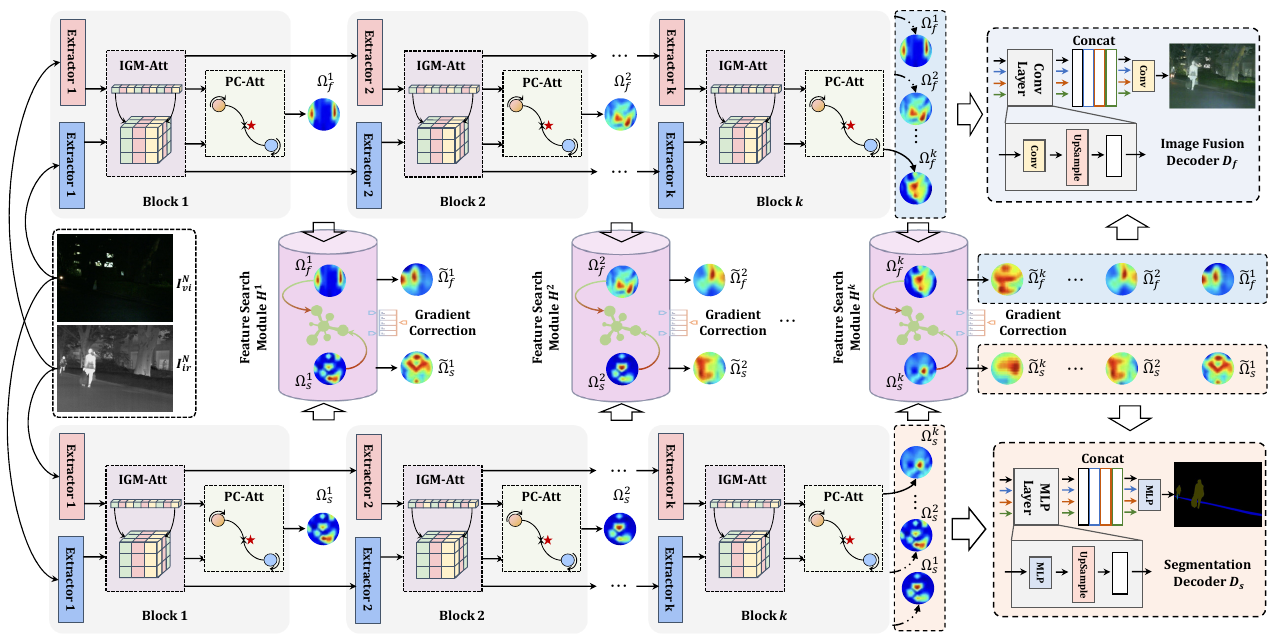}
	\caption{Overall framework of our W2F for mutually reinforcing visual fusion and segmentation via visual-semantic collaboration.}\label{fig2}
\end{figure*}

\section{Methodology}\label{sec:methods}
W2F achieves a dual performance boost by mutually reinforcing visual fusion and semantic segmentation. In this section, we provide the problem formulation, and present the complementarity strengthening strategy, the multi-modal information interaction mechanism, and the collaborative optimization mechanism in detail.

\subsection{Problem Formulation}
Our goal is to seamlessly integrate visual fusion and semantic segmentation to reduce the optimization gap between the visual and semantic domains in the perception enhancement system. To this end, we propose W2F to establish a mutual reinforcement paradigm between visual fusion and segmentation via visual-semantic collaboration, as shown in Fig.~\ref{fig2}.

Now, let us formally define this research problem. Given a pair of degraded multi-modal images $\{I^{{N}}_{vi}, I^{N}_{ir}\}$ captured in harsh environments, our perception enhancement system aims to obtain a comprehensive representation of the imaging scene. Here, $vi$ and $ir$ denote the visible and infrared modalities, respectively, and $N$ indicates the composite degradation. This goal involves generating fused images $I_{f}$ with promising visual perception effects and segmentation results $I_{s}$ with high decision accuracy:
\begin{equation}	
	I_{f}= F(I^{{N}}_{vi}, I^{{N}}_{ir}), \quad  I_{s}= S(I^{{N}}_{vi}, I^{{N}}_{ir}),\label{Eq1}
\end{equation}
where $F$ and $S$ represent the visual fusion network and the segmentation network, respectively. Both are composed of isomorphic encoders and their respective decoders. Therefore, Eq.~\eqref{Eq1} can be further specified as:
\vspace{-0.4cm}	
\begin{equation}
	I_{f}= D_f(E_f(I^{{N}}_{vi}, I^{{N}}_{ir})), \quad  I_{s}= D_s(E_s(I^{{N}}_{vi}, I^{{N}}_{ir})).\label{Eq2}
\end{equation}

To achieve mutual reinforcement between visual fusion and semantic segmentation, we use encoders $E_f$ and $E_s$ as a bridge to establish the visual-semantic collaboration mechanism. Before this, two prerequisites need to be met. First, the bidirectional representational complementarity between visual fusion and semantic segmentation should be sufficiently strengthened to establish a reliable basis for their subsequent collaboration. To this end, we design a segmentation-guided fusion loss to inject object-aware structural guidance into visual fusion, improving object visibility and distinguishability. Meanwhile, an information restoration prior, covering multiple degradation types, is embedded into the optimization objective of visual fusion. This strengthens its capability to extract reliable visual perceptual evidence from degraded observations $\{I^{{N}}_{vi}, I^{N}_{ir}\}$, thereby enhancing its ability to assist the semantic branch in constructing reliable object-level representations. Together, these two designs strengthen the bidirectional representational complementarity between visual fusion and semantic segmentation, providing a solid basis for their subsequent collaboration. Second, serving as the bridge for visual-semantic collaboration, the encoders should possess strong multi-modal interaction capabilities to provide solid information support for the collaboration. Therefore, we develop two powerful attention modules to strengthen the mutual completion and aggregation of multi-modal features in the encoders $E_f$ and $E_s$, \textit{i.e.}, IGM-Att and PC-Att. Specifically, based on the CNN paradigm, IGM-Att leverages statistical pooling to assess the importance of each modality's features at both the channel and spatial levels, and further introduces a trainable gating mechanism to correct potential mismatched scores. According to the corrected scores, multi-modal features can selectively extract the necessary information from other-modal features to locally refine themselves. Furthermore, PC-Att couples self-attention and cross-attention, progressively achieving single-modal self-reinforcement and cross-modal mutual complementarity. Finally, it globally aggregates multi-modal features, laying the informational foundation for comprehensive scene perception.

Building upon the above foundations, we further utilize the interaction between encoders to achieve the desired mutual reinforcement between visual fusion and semantic segmentation. First, a feature search module $\varrho$ is developed to enable these two tasks to retrieve complementary information from each other’s encoded features, thereby enhancing their visual and semantic representations through visual-semantic collaboration. Formally, the encoded features in visual fusion and semantic segmentation are denoted as $\Omega_f=E_f(I^{{N}}_{vi}, I^{{N}}_{ir})$ and $\Omega_s=E_s(I^{{N}}_{vi}, I^{{N}}_{ir})$, respectively. The enhanced encoded features produced by the feature search module are formulated as:
\begin{equation}	
	\{\tilde{\Omega}_f, \tilde{\Omega}_s\} = \varrho(\Omega_f,\Omega_s) .\label{Eq3}
\end{equation}

Notably, although $\Omega_f$ and $\Omega_s$ both originate from $\{I^{{N}}_{vi}, I^{N}_{ir}\}$, their representations gradually become task-biased under the respective optimization objectives of visual fusion and semantic segmentation. Through feature search, complementary visual and semantic information can be effectively injected into each other, thereby enhancing the representational capability of both tasks and enabling further performance improvement. Based on the enhanced features, the fused image $\tilde{I}_{f}$ with higher visual quality and the segmentation result $\tilde{I}_{s}$ with higher semantic accuracy can be obtained:
\begin{equation}	
	\tilde{I}_{f}= D_f(\tilde{\Omega}_f), \quad  \tilde{I}_{s}= D_s(\tilde{\Omega}_s).\label{Eq4}
\end{equation}

However, due to parameter reuse in the feature search module $\varrho$, potential optimization conflicts may arise between the visual fusion and semantic segmentation. To address this issue, we employ a gradient correction strategy to ensure collaborative consistency between these two tasks from the optimization perspective. Specifically, for the visual fusion loss $\mathcal{L}_F$ and the semantic segmentation loss $\mathcal{L}_S$, the gradients with respect to the reused parameters can be represented as:
\begin{equation}	
	\delta_f= \frac{\partial \mathcal{L}_F}{\partial\sigma}, \quad  \delta_s= \frac{\partial \mathcal{L}_S}{\partial \sigma}.\label{Eq5}
\end{equation}
where $\partial$ represents the partial derivative, $\sigma$ denotes the reused parameters, $\delta_f$ and $\delta_s$ denote the corresponding optimization gradients. When the angle between $\delta_f$ and $\delta_s$ is obtuse ($\angle(\delta_f,\delta_s) > 90^\circ$), an optimization conflict exists between the two tasks. In this case, we randomly select the gradient of one task and project it onto the normal direction of the other task's gradient to eliminate the conflict~\cite{yu2020gradient}. Furthermore, we aim to maintain synchronization between the visual fusion and segmentation tasks as much as possible for all non-conflicting cases ($\angle(\delta_f,\delta_s) \leq 90^\circ$). Therefore, Nash-MTL~\cite{navon2022multi} is adopted to produce a Nash bargaining solution for these two tasks by combining per-task gradients into a joint update direction. With the above safeguards, W2F establishes an effective mutual reinforcement between visual fusion and segmentation, enabling comprehensive perception enhancement from visual to semantic levels.

\subsection{Complementarity Amplification for Collaboration}
\begin{figure}[t]
	\centering
	\includegraphics[width=1\linewidth]{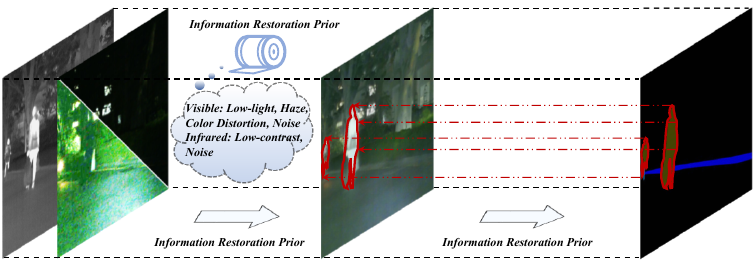}
	\vspace{-0.15in}
	\caption{Illustration of amplifying the complementarity between visual fusion and semantic segmentation, which strengthens the basis for their collaboration.}\label{fig3}
\end{figure}

To amplify the complementarity between visual fusion and semantic segmentation, our W2F introduces two targeted designs, namely segmentation-guided fusion loss and information restoration prior embedding, as shown in Fig.~\ref{fig3}.

\subsubsection{Segmentation-Guided Fusion Loss}
The segmentation-guided fusion loss aims to explicitly inject semantic-aware structural priors into the visual fusion optimization, thereby strengthening the guidance of semantic representations for fused image generation. Since segmentation focuses more on object contours, foreground saliency, and category-discriminative cues, its encoded features can guide the visual fusion features to pay more attention to semantically important regions, thereby improving object visibility, edge clarity, and foreground distinguishability in fused images. Formally, according to Eq.~\eqref{Eq1}, given the degraded source images $\{I^{N}_{vi}, I^{N}_{ir}\}$, the fusion network $F$ generates the fused image $I_f$. We use pixel intensity as the foundational representation of images and construct the segmentation-guided intensity loss:
\begin{equation}
	\mathcal{L}^{SI}_{F}\!(I_f)\!=\!\!\left\{
	\begin{array}{ll}
		\!\!{\left\|T\!\cdot\!(I_{f}\!-\!\max(I^{N}_{vi}, I^{N}_{ir}))\right\|}_1, \!\vspace{1ex} &\!\!\!\!\text{if }\!T \!\in\! \arg_\tau\! (\!\ddot{I}_s\!), \\
		\!\!\!\sum\limits_{i}^{\{vi, ir\}} \!\omega_{i} \cdot \!{\left\|T\cdot(I_{f}- I^{N}_{i}))\right\|}_1, &\!\! \text{otherwise},
	\end{array} \right.\label{Eq6}
\end{equation}
where $\arg$ is a search operator used to locate pixels belonging to the specified semantic category set $\tau$ in segmentation labels $\ddot{I}_s$. Importantly, $\tau$ is not restricted to fixed categories, but is determined according to application requirements. In traffic perception scenarios, thermal-salient categories such as ``person'' and ``car'' are typically included because they correspond to critical foreground objects. More generally, $\tau$ can be flexibly extended to other task-relevant categories, such as ``bus'' or ``truck'', depending on the target application and annotation setting. In this work, $\tau$ is specified as ``person'' and ``car''. $T$ denotes the screening map used to provide different fusion guidance for different semantic regions.

For regions belonging to $\tau$, the $\max$ function is adopted to preserve the most significant intensity responses from multi-modal inputs. Especially for thermal-salient objects such as persons and cars, which usually exhibit high intensity in infrared images, their saliency can be effectively preserved through the maximum operation. In this way, the fused image is encouraged to highlight semantically important foreground objects while maintaining clearer structural visibility. In contrast, for non-$\tau$ regions, dynamic weighting guidance is introduced based on regional information effectiveness evaluation. Specifically, local contrast statistics from multiple image patches are aggregated to measure the information richness of each modality, based on the observation that images with stronger local contrast generally contain clearer structural details. Thus, the informational effectiveness score of a single-modal image is defined as $\varepsilon=\sum_p{SD(p)}$, where $p$ denotes the image patch and $SD$ represents the standard deviation function. Accordingly, the weighting factor in Eq.~\eqref{Eq6} is defined as $\omega_{i}=\varepsilon_{i}/\sum_{i} \varepsilon_{i}$, where $i$ denotes the modality type. Through Eq.~\eqref{Eq6}, semantic-aware structural constraints are explicitly incorporated into visual fusion optimization, enabling semantic features to become deeply involved in fused image generation and thereby improving the visual representation capability of fused images. It strengthens the unidirectional complementarity from segmentation to visual fusion, providing important support for visual-semantic collaboration.

\subsubsection{Information Restoration Prior and Expert Dataset}
Compared with the semantic-to-visual complementarity, the benefit of visual fusion for semantic segmentation is less intuitive. To strengthen this complementarity, we advocate embedding information restoration prior into the optimization objective of visual fusion. The core motivation is that semantic segmentation usually struggles to perform effectively on degraded multi-modal data captured in harsh environments. If the visual fusion network $F$ can simultaneously learn degradation removal and information fusion, it can extract more reliable visual perceptual evidence from multi-modal degraded images $\{I^{N}_{vi}, I^{N}_{ir}\}$, thereby strengthening its ability to assist the semantic segmentation model $S$ in constructing reliable object-level representations. Importantly, the embedding of information restoration prior must rely on visual-level supervision. Specifically, degradation removal capability should first be learned through constraints imposed on fused image reconstruction, and can then be transferred to the semantic branch through visual-semantic collaboration. Therefore, instead of directly introducing restoration constraints into semantic segmentation, W2F embeds the restoration prior into visual fusion optimization.

In our W2F, the embedding of information restoration prior is achieved through a modification of Eq.~\eqref{Eq6}. More concretely, we break from the conventional idea that constructs visual fusion loss based on the similarity between the fused image and source images, and instead introduce clean images as the reconstruction target. Thus, Eq.~\eqref{Eq6} can be rewritten as:
\begin{equation}
	\mathcal{L}^{SI}_{F}\!(I_f)\!=\!\!\left\{
	\begin{array}{ll}
		\!\!{\left\|T\!\cdot\!(I_{f}\!-\! \max(I^{E}_{vi}, I^{E}_{ir}))\right\|}_1,\! \vspace{1ex} & \text{if }\!T \!\in\! \arg_\tau\! (\!\ddot{I}_s\!), \\
		\!\!\!\sum\limits_{i}^{\{vi, ir\}}\!\!\omega_{i}\! \cdot \!{\left\|T\!\cdot\!(I_{f}\!-\! I^{E}_{i}))\right\|}_1, &\!\!  \text{otherwise},
	\end{array} \right.\label{Eq7}
\end{equation}
where $I^{E}_{vi}$ and $I^{E}_{ir}$ represent the clean images obtained after refining $I^{N}_{vi}$ and $I^{N}_{ir}$, respectively.

\begin{figure}[t]
	\centering
	\includegraphics[width=1\linewidth]{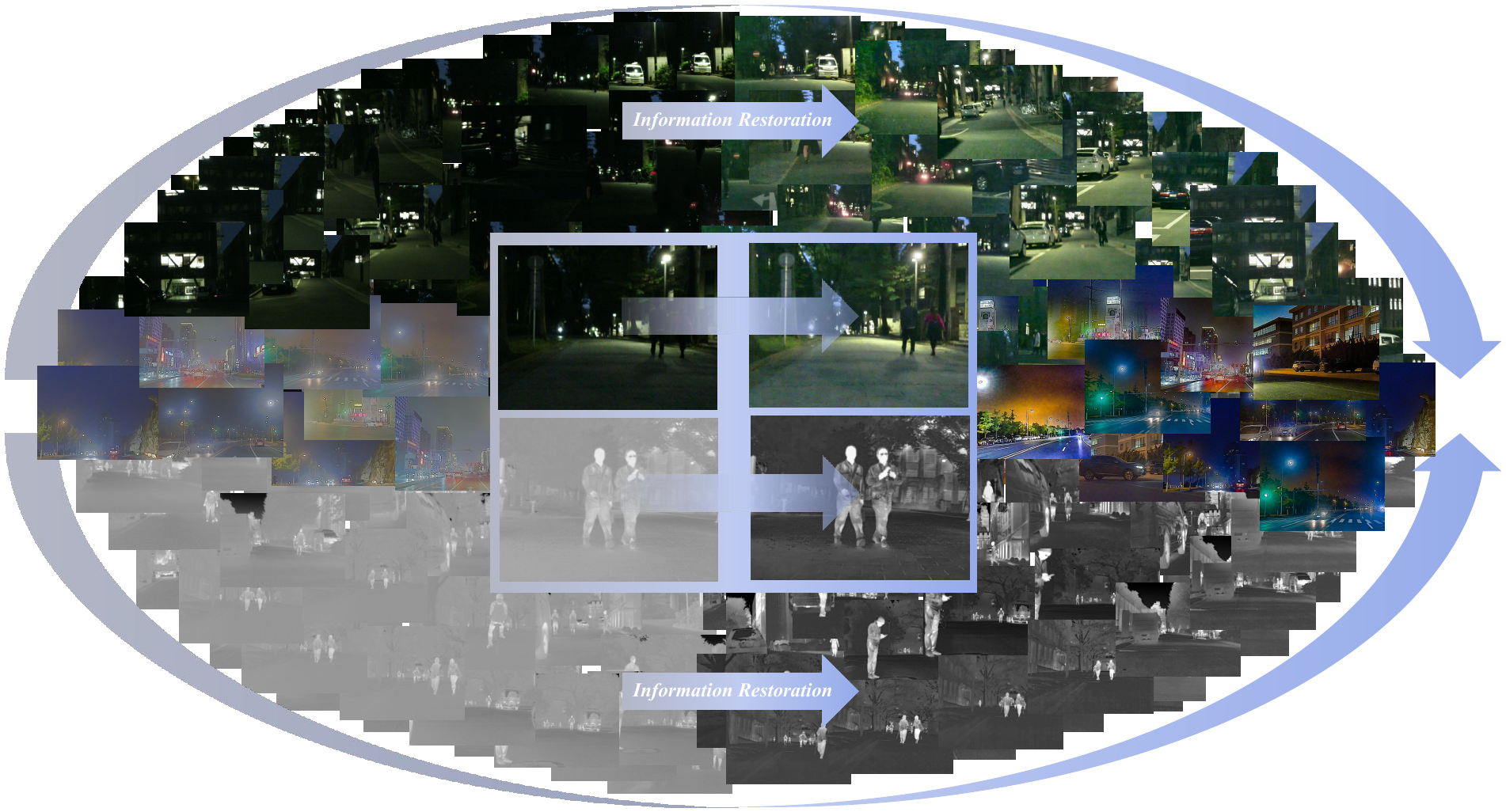}
	\caption{New high-quality datasets constructed using Adobe Photoshop Lightroom, containing $1,569$ and $1,500$ spatially aligned clean visible and infrared reference image pairs for MFNet~\cite{ha2017mfnet} and FMB~\cite{liu2023multi}, respectively, to provide the desired information restoration prior.}\label{fig4}
\end{figure}

To ensure the effectiveness of the information restoration prior, we invite $10$ visual experts to manually construct refined clean reference images $I^{E}_{vi}$ and $I^{E}_{ir}$ based on the MFNet~\cite{ha2017mfnet} and FMB~\cite{liu2023multi}. MFNet and FMB contain $1,569$ and $1,500$ original scenes, respectively, with each scene consisting of one visible image and one spatially aligned infrared image. For each scene, the $10$ experts independently restore both modalities using Adobe Photoshop Lightroom~\cite{lightroom2012adobe}, producing $10$ candidate restorations for each modality. Specifically, visible image restoration mainly focuses on enhancing low-light visibility, suppressing imaging noise, correcting color distortion, and reducing haze, whereas infrared image restoration mainly focuses on improving image contrast and suppressing imaging noise. This process yields $15,690$ visible and $15,690$ infrared candidate restorations for MFNet, and $15,000$ visible and $15,000$ infrared candidates for FMB.

We further conduct modality-wise cross-expert evaluation and ranking for quality control. For each scene and modality, each expert evaluates only the $9$ candidate restorations generated by the other $9$ experts, without evaluating their own restoration, and ranks them from $1$ (best) to $9$ (worst). The evaluation considers whether major degradations are effectively alleviated, whether the original scene structures, object boundaries, and semantic contents are faithfully preserved, and whether artifacts, over-enhancement, or unrealistic colors are introduced. Consequently, each candidate receives $9$ rankings, whose mean serves as its quality score, with a lower mean rank indicating better restoration quality. For each modality, the candidate with the lowest mean rank is selected as the reference image. In the case of a tie, one of the tied candidates is randomly selected. The selected visible and infrared reference images are then recombined according to their original spatial correspondence to form the final clean reference pair. Overall, this quality-control process involves $141,210$ visible-image evaluations and $141,210$ infrared-image evaluations for MFNet, and $135,000$ visible-image evaluations and $135,000$ infrared-image evaluations for FMB. After the evaluation and selection process, we obtain $1,569$ spatially aligned clean visible and infrared reference image pairs for MFNet and $1,500$ for FMB, with each pair corresponding to one original scene, as shown in Fig.~\ref{fig4}. With the introduced restoration prior and expert-refined datasets, W2F enables the visual fusion branch to learn higher-visibility scene representation, thereby enhancing its ability to support semantic reasoning under degraded conditions.

\begin{figure}[t]
	\centering
	\includegraphics[width=1\linewidth]{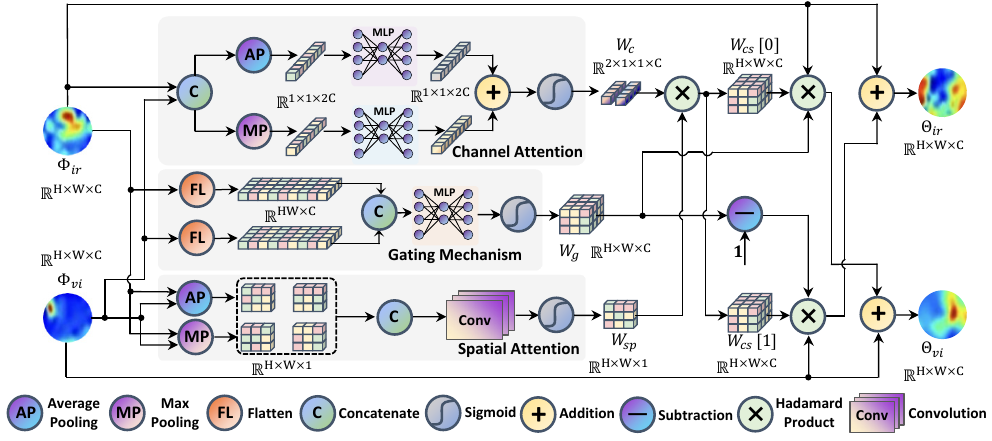}
	\caption{Architecture of our proposed IGM-Att module.}\label{fig5}
\end{figure}

\subsection{Multi-Modal Information Interaction}
An effective multi-modal information interaction mechanism forms the information foundation for visual-semantic collaboration. In our W2F, multi-modal information interaction occurs within the encoders, as shown in Fig.~\ref{fig2}. Specifically, the encoder continuously extracts and refines multi-modal features through $k$ blocks containing IGM-Att and PC-Att modules. In each block, the encoder initially conducts feature extraction:
\vspace{-0.4cm}
\begin{equation}	
	\Phi_{vi}^{k}=U_{vi}^{k}(\Theta_{vi}^{k-1}), \quad  \Phi_{ir}^{k}=U_{ir}^{k}(\Theta_{ir}^{k-1}),\label{Eq8}
\end{equation}
where $U_{vi}^{k}$ and $U_{ir}^{k}$ denote $k$-th feature extraction units on the visible and infrared branches, respectively. $\{\Theta_{vi}^{k-1}$, $\Theta_{ir}^{k-1}$\} represent the visible and infrared features refined by the $(k-1)$-th IGM-Att module. These features are initialized by the input degraded visible and infrared images $\{I^{N}_{vi}, I^{N}_{ir}\}$ when $k=1$. Then, we design an IGM-Att module to facilitate interaction between $\Phi_{vi}^{k}$ and $\Phi_{ir}^{k}$. This module employs the CNN paradigm to predominantly emphasize local mutual completion of cross-modal features. Besides, it incorporates a gating mechanism to rectify misjudgments in the conventional pooling-based attention. Thus, the mutually reinforced visible and infrared features are obtained by:
\begin{equation}	
	\{\Theta_{vi}^{k},\Theta_{ir}^{k}\}=M^{k}(\Phi_{vi}^{k}, \Phi_{ir}^{k}),\label{Eq9}
\end{equation}
where $M^{k}$ denotes the function of the $k$-th IGM-Att. On this basis, PC-Att is developed to further facilitate interaction and aggregation of multi-modal features. In contrast to IGM-Att, PC-Att places greater emphasis on the global integration of multi-modal features for scene understanding. Consequently, we establish a feature progressive transfer and fusion architecture within the Transformer paradigm. The aggregated features obtained after PC-Att processing can be expressed as:
\begin{equation}	
	\{\Omega_f^{k}/\Omega_s^{k}\}=P^{k}(\Theta_{vi}^{k}, \Theta_{ir}^{k}),\label{Eq10}
\end{equation}
where $P^{k}$ represents the function of the $k$-th PC-Att. Due to the comprehensive consideration of single-modal self-reinforcement and cross-modal mutual complementarity, the aggregated features $\Omega_f^{k}$ and $\Omega_s^{k}$ encompass complete information describing the scene. Thus, they can provide strong informational support for both visual fusion and semantic segmentation. Next, we provide detailed descriptions of IGM-Att and PC-Att.

\begin{figure}[t]
	\centering
	\includegraphics[width=0.90\linewidth]{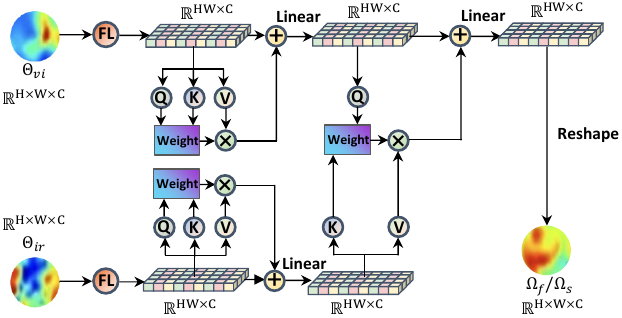}
	\caption{Architecture of our proposed PC-Att module.}\label{fig6}
\end{figure}

\subsubsection{IGM-Att}
The IGM-Att module integrates a gating mechanism into the conventional pooling-based attention for feature completion, as shown in Fig.~\ref{fig5}. First, it adopts conventional channel and spatial attention~\cite{woo2018cbam} to model the relative relationship between multi-modal features. In the channel attention, $\Phi_{vi}^{k}\in \mathbb{R}^{H\times W \times C}$ and $\Phi_{ir}^{k}\in \mathbb{R}^{H\times W \times C}$ are concatenated along the channel dimension, and then reduced by the max and average pooling to capture the response properties of each channel. Next, we use MLP to project these maximum and average responses separately. The resulting vectors are summed and activated by the Sigmoid function to generate channel-attention weights, denoted as $W_{c}^{k}\in \mathbb{R}^{2\times 1\times 1 \times C}$. This process can be formalized as:
\begin{equation}
	\begin{aligned}
		W_{c}^{k}= & Sig( MLP(AP(C(\Phi_{vi}^{k},\Phi_{ir}^{k})))\\
		&+MLP(MP(C(\Phi_{vi}^{k},\Phi_{ir}^{k})))).
	\end{aligned}	\label{Eq11}
\end{equation}
In the spatial attention, we use max and average pooling on both $\Phi_{vi}^{k}$ and $\Phi_{ir}^{k}$, yielding four maps that capture their pixel-wise response properties. These maps are concatenated along the channel dimension, and then processed by two convolutions with the Sigmoid function to produce the spatial-attention weights, denoted as $W_{sp}^{k}\in \mathbb{R}^{H\times W \times 1}$. We formalize it as:
\begin{equation}	
	W_{sp}^{k} \!=\! Sig(Conv(C(AP(\Phi_{vi}^{k},\!\Phi_{ir}^{k}),\! MP(\Phi_{vi}^{k},\!\Phi_{ir}^{k})))).\label{Eq12}
\end{equation}
We derive hybrid weights $W_{cs}^{k} \in \mathbb{R}^{2\times H\times W \times C}$, by multiplying channel-attention weights with spatial-attention weights, expressed as $W_{cs}^{k} = W_{c}^{k} \times W_{sp}^{k}$. Nevertheless, as discussed in MRFS~\cite{zhang2024mrfs}, such pooling-based attention may lead to mismatched feature validity scores. To address this, we introduce a gating mechanism to correct the obtained hybrid weights $W_{cs}^{k}$. Specifically, we first flatten $\Phi_{vi}^{k}$ and $\Phi_{ir}^{k}$. The flattened features are then concatenated along the channel dimension and input into an MLP with a Sigmoid function to derive the gated correction scores $W_{g}^{k} \in \mathbb{R}^{H\times W \times C}$. We encapsulate the entire gating mechanism process as:
\begin{equation}	
	W_{g}^{k} = Sig(MLP(C(FL(\Phi_{vi}^{k}), FL(\Phi_{ir}^{k})))).\label{Eq13}
\end{equation}
Next, the hybrid weights undergo refinement using the correction scores:
\begin{equation}
	\left\{
	\begin{aligned}
		&W_{csg-vi}^{k} = W_{cs}^{k}[0]\times W_{g}^{k},  \\
		&W_{csg-ir}^{k} = W_{cs}^{k}[1]\times (1-W_{g}^{k}) .
	\end{aligned}	\label{Eq14}
	\right.
\end{equation}
This sophisticated strategy enhances the precision of feature importance assessments, reinforcing valuable features and effectively suppressing redundant ones. Finally, we refine features for local completion using the obtained weights:
\begin{equation}
	\left\{
	\begin{aligned}
		&\Theta_{vi}^{k} = \Phi_{vi}^{k} + W_{csg-vi}^{k} \times \Phi_{ir}^{k},  \\
		&\Theta_{ir}^{k} = \Phi_{ir}^{k} + W_{csg-ir}^{k} \times \Phi_{vi}^{k} .
	\end{aligned}	\label{Eq15}
	\right.
\end{equation}

\subsubsection{PC-Att}
The IGM-Att module leverages CNN-based attention, emphasizing the refinement of local features. However, in further information integration, a global understanding of the scene is required. Therefore, we develop a PC-Att module to achieve this goal, illustrated in Fig.~\ref{fig6}. In the PC-Att module, we employ two information refinement strategies, namely, single-modal self-reinforcement and cross-modal mutual complementarity. For single-modal self-reinforcement, we compute Query ($Q_X\in \mathbb{R}^{HW\times C}$), Key ($K_X\in \mathbb{R}^{HW\times C}$), and Value ($V_X\in \mathbb{R}^{HW\times C}$) using linear layers from the feature $\Upsilon_X \in \mathbb{R}^{HW\times C}$. $Q^X$ is then employed to query $K_X$ for allocation scores, and utilize these scores to modulate $V_X$ to obtain the refined value:
\begin{equation}	
	{V_X^{'}} = softmax\bigg(\frac{Q_X {K_X^T}}{\sqrt{d_k}}\bigg)V_X.\label{Eq16}
\end{equation}
Subsequently, the refined value ${V_X^{'}}$ is used to enhance $\Upsilon_X$, resulting in improved features:
\begin{equation}	
	{\Upsilon_X^{self}} = Linear(\Upsilon_X + {V_X^{'}}).\label{Eq17}
\end{equation}
This entire process is formally expressed as ${\Upsilon_X^{self}} = Self(\Upsilon_X)$. With this design, valuable cues from the single-modal image itself (\textit{e.g.}, objects of the same class) can be leveraged for feature refinement. However, the ability of self-reinforcement is inherently limited, so we introduce the cross-modal mutual complementarity. Without loss of generality, we describe its process by using the X-modal feature $\Upsilon_X$ to enhance the Y-modal feature $\Upsilon_Y$. $Q_Y$ is derived from $\Upsilon_Y$ to represent the demand, while $K_X$ and $V_X$ are extracted from $\Upsilon_X$. Following a procedure similar to self-reinforcement, we leverage $Q_Y$ and $K_X$ to compute the allocation score and modulate $V_X$:
\begin{equation}	
	{V_{Y\gets X}} = softmax\bigg(\frac{Q_Y {K_X}^{T}}{\sqrt{d_k}}\bigg)V_X.\label{Eq18}
\end{equation}
The cross-modal refined feature is obtained by:
\begin{equation}	
	{\Upsilon_Y^{cross}} = Linear(\Upsilon_Y + {V_{Y\gets X}}).\label{Eq19}
\end{equation}
This cross-modal mutual complementarity process is formally expressed as ${\Upsilon_Y^{cross}} \!=\! Cross(\Upsilon_Y,\Upsilon_X)$. Referring to Fig.~\ref{fig6}, in our PC-Att module, we flatten $\Theta_{vi}^{k}$ and $\Theta_{ir}^{k}$ and apply the self-reinforcement and cross-modal complementarity strategies in a progressive manner:
\begin{equation}
	\left\{
	\begin{aligned}
		&{\Upsilon_{vi}^{k}}^{self} = Self(FL(\Theta_{vi}^{k})),  \\
		&{\Upsilon_{ir}^{k}}^{self} = Self(FL(\Theta_{ir}^{k})),  \\
		&\{\Omega_f^{k}/\Omega_s^{k}\} \!=\!  Cross\big(FL\big({\Upsilon_{vi}^{k}}^{\!self\!}\big),{\Upsilon_{ir}^{k}}^{\!self\!}\big).
	\end{aligned}	\label{Eq20}
	\right.
\end{equation}
Such a progressive process seamlessly integrates long-distance complete scene information from infrared and visible images into the fused features $\Omega_{f}^{k}$ and $\Omega_{s}^{k}$.

\begin{figure}[t]
	\centering
	\includegraphics[width=1\linewidth]{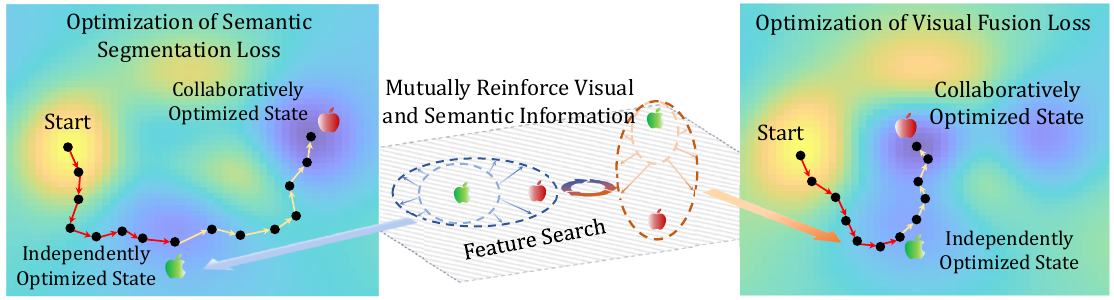}
	\caption{Optimization illustration of the proposed feature search module. The two curved surfaces denote the optimization landscapes of the visual fusion and semantic segmentation losses, respectively. The red trajectories represent the independent optimization processes leading to the \textit{Independently Optimized State}, where further performance improvement is constrained by the lack of cross-task complementarity. The white trajectories represent the subsequent collaborative optimization that progressively exploits complementary cross-task representations to break this performance bottleneck, leading to the \textit{Collaboratively Optimized State}.}\label{fig7}
\end{figure}

\subsection{Collaborative Optimization}
Now that the visual-semantic complementarity has been amplified and the multi-modal information interaction mechanism has been established, we further construct a collaborative optimization mechanism between visual fusion and semantic segmentation. It consists of a feature search module and a gradient correction strategy, which jointly ensure effective visual-semantic mutual reinforcement from the perspectives of feature interaction and optimization consistency, respectively.

\subsubsection{Feature Search Module}
Our W2F adopts a two-stage optimization strategy. In the first stage, the visual fusion network $F$ and the semantic segmentation network $S$ are independently optimized with their respective losses, $\mathcal{L}_{F}$ and $\mathcal{L}_{S}$. Naturally, the encoded features $\Omega_{f}$ and $\Omega_{s}$ gradually become task-specific representations. Specifically, $\Omega_{f}$ tends to focus more on visually favorable texture details and intensity contrast representations, while $\Omega_{s}$ emphasizes semantic-aware structural cues and category-discriminative representations. However, since the two tasks are independently optimized, each task relies solely on its own representations without explicitly exploiting the complementary representations from the other task, thereby constraining further performance improvement and leading to a performance bottleneck. We refer to this stage as the \textit{Independently Optimized State}, as illustrated by the red trajectories in Fig.~\ref{fig7}.

\begin{figure}[t]
	\centering
	\includegraphics[width=1\linewidth]{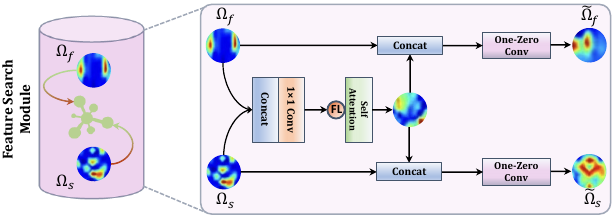}
	\caption{Architecture of our proposed feature search module.}\label{fig8}
\end{figure}

In the second stage, the independently trained encoders from the visual fusion and the semantic segmentation branches are frozen. On this basis, we introduce a feature search module between the visual and semantic encoder blocks at each stage, enabling $\Omega^{k}_{f}$ and $\Omega^{k}_{s}$ to preserve their own task-specific representations while retrieving complementary information from each other for further enhancement.  In this way, both visual and semantic features can continuously enrich themselves without losing their original representational characteristics. The detailed structure of the feature search module is illustrated in Fig.~\ref{fig8}. First, $\Omega^{k}_{f}$ and $\Omega^{k}_{s}$ are fully integrated to construct a shared feature pool:
\begin{equation}	
	\Omega^{k}_{f-s} = Self(FL(Conv(C(\Omega^{k}_{f}, \Omega^{k}_{s})))).\label{Eq21}
\end{equation}
where concatenation and convolution are employed to linearly combine visual and semantic features, followed by a self-attention operation for global feature refinement. The obtained shared feature pool simultaneously contains visual-oriented and semantic-oriented representations, providing a unified space for subsequent visual-semantic feature retrieval. Then, $\Omega^{k}_{f}$ and $\Omega^{k}_{s}$ selectively retrieve the complementary information they require from the shared feature pool to further enhance themselves:
\begin{equation}
	\left\{
	\begin{aligned}
		&\tilde{\Omega}_f^k = Conv_{1-0}(C(\Omega^{k}_{f},\Omega^{k}_{f-s})),  \\
		&\tilde{\Omega}_s^k = Conv_{1-0}(C(\Omega^{k}_{s},\Omega^{k}_{f-s})),
	\end{aligned}	\label{Eq22}
	\right.
\end{equation}
where $Conv_{1-0}$ denotes One-Zero convolution, whose convolution kernels are initialized in a binary manner at the beginning of optimization. Inspired by the zero convolution design in ControlNet~\cite{zhang2023adding}, this strategy enables the enhanced features, $\tilde{\Omega}_f^k$ and $\tilde{\Omega}_s^k$, to be progressively learned based on the original task-specific representations  $\Omega^{k}_{f}$ and $\Omega^{k}_{s}$, thereby avoiding unstable feature interaction during the early training stage and preserving the optimization foundations already learned by each task. Based on the enhanced features $\tilde{\Omega}_f^k$ and $\tilde{\Omega}_s^k$, the visual fusion decoder $D_f$ and segmentation decoder $D_s$ can further generate fused images with clearer structural details and segmentation results with higher semantic accuracy, respectively.

Since the second-stage collaborative optimization fully inherits the task-specific representations learned in the first stage, and the feature search module is initialized with One-Zero convolution, the collaborative optimization naturally starts from the \textit{Independently Optimized State} (\textit{i.e.}, green apples in Fig.~\ref{fig7}). Building upon these preserved task-specific representations, the progressively injected complementary information enables each task to exploit beneficial representations from the other task, thereby breaking the performance bottleneck caused by independent optimization. We refer to the resulting stage as the \textit{Collaboratively Optimized State}, as illustrated by the white trajectories in Fig.~\ref{fig7}.

\subsubsection{Gradient Correction Strategy}
Although the feature search module establishes feature-level collaboration between the two tasks, jointly optimizing visual fusion and semantic segmentation may still introduce optimization conflicts during training. To further ensure effective visual-semantic collaboration from the optimization perspective, we adopt a gradient correction strategy to coordinate the optimization behaviors of the visual fusion loss $\mathcal{L}_{F}$ and the semantic segmentation loss $\mathcal{L}_{S}$. Notably, the proposed gradient correction strategy mainly operates on the feature search modules. This is because the independently trained visual fusion and semantic segmentation branches are frozen during the second stage, while parameter reuse only occurs within the feature search modules introduced for visual-semantic collaboration. Therefore, potential optimization conflicts between visual fusion and semantic segmentation are also concentrated in these shared modules.

Specifically, according to PCGrad~\cite{yu2020gradient}, when the angle between the gradients of visual fusion and semantic segmentation is obtuse, \textit{i.e.}, $\angle(\delta_f,\delta_s) > 90^\circ$, the optimization directions of the two tasks become conflicting. At this point, we randomly select the gradient of one task and project it onto the normal plane of the other task's gradient:
\begin{equation}	
	\delta_i \leftarrow \delta_i-\frac{\delta_i \cdot \delta_j}{\left\|\delta_j\right\|^2}\delta_j, \quad \{i\neq j\} \in rand(f,s). \label{Eq23}
\end{equation}
In this way, the conflicting gradient component is removed, significantly reducing the risk of optimization interference between visual fusion and semantic segmentation. After correction, the optimization relationship between the two tasks becomes either mutually beneficial or non-conflicting, \textit{i.e.}, \(\angle(\delta_f, \delta_s) \leq 90^\circ\).

\begin{figure}[t]
	\centering
	\includegraphics[width=0.98\linewidth]{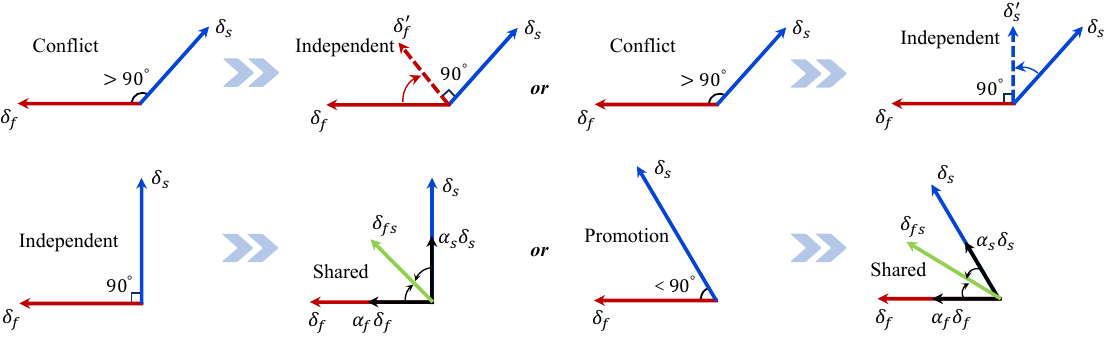}
	\caption{Illustrating the gradient correction strategy.}\label{fig9}
\end{figure}

However, eliminating gradient conflicts alone is insufficient for achieving stable visual-semantic collaboration. Even under non-conflicting optimization directions, inconsistent learning progress between visual fusion and semantic segmentation may still lead to optimization imbalance. In such cases, one task may dominate the collaborative optimization process while the other fails to sufficiently learn complementary representations, thereby weakening the effectiveness of visual-semantic collaboration. To address this issue, we further employ NASH-MTL~\cite{navon2022multi} to balance the collaborative optimization between visual fusion and semantic segmentation through a bargaining mechanism. Specifically, NASH-MTL seeks a Nash Bargaining Solution to achieve proportionally balanced optimization benefits for both tasks. Thus, a shared collaborative gradient can be obtained:
\begin{equation}	
	\delta_{fs} = \sum\limits_{i}\alpha_{i} \cdot \delta_{i}, \quad i \in {f,s}. \label{Eq24}
\end{equation}
The coefficient $\alpha$ used for aggregating gradients is obtained by solving $GG^\top\alpha=1/\alpha$, where $G$ is the gradient matrix formed by concatenating $\delta_f$ and $\delta_s$ column-wise. In this way, visual fusion and semantic segmentation jointly update the shared feature search modules using the collaborative gradient $\delta_{fs}$, thereby achieving more consistent optimization. Overall, the employed gradient correction strategy effectively coordinates the optimization behaviors of visual fusion and semantic segmentation within the shared feature search modules. The overall process of the proposed gradient correction strategy is illustrated in Fig.~\ref{fig9}.

\subsection{Loss Functions}
Now, we provide the complete definitions of visual fusion loss $\mathcal{L}_{F}$ and semantic segmentation loss $\mathcal{L}_{S}$.

\subsubsection{Visual Fusion Loss}
The segmentation-guided intensity loss $\mathcal{L}^{SI}_{F}$ in Eq.~\eqref{Eq7} mainly constrains semantic-aware intensity preservation and foreground object saliency in the fused image. However, texture structures and color fidelity still require further constraints. Therefore, we further introduce a texture integration loss:
\begin{equation}
	\mathcal{L}^{TI}_{F}(I_{f})={\left\|\nabla I_{f}- \max(\nabla I^{E}_{vi}, \nabla I^{E}_{ir})\right\|}_1,\label{Eq25}
\end{equation}
where $\nabla$ denotes the Sobel gradient operator. Intuitively, gradients are used to characterize texture structures, and the loss encourages the fused image to preserve the most significant gradients, thereby generating clearer structural and texture representations. Besides, a color fidelity loss is defined as:
\begin{equation}
	\mathcal{L}^{CF}_{F}(I_{f})={\left\|CbCr(I_{f})- CbCr(I^{E}_{vi})\right\|}_1,\label{Eq26}
\end{equation}
where $CbCr$ refers to the operation of extracting the chrominance components after converting the RGB image into the YCbCr color space. It encourages the fused image to maintain color consistency with the visible image, thereby producing results that better conform to human visual perception. In summary, the visual fusion loss can be formally constructed as:
\begin{equation}
	\mathcal{L}_{F}(I_{f})=\mathcal{L}^{SI}_{F}(I_{f})+\mathcal{L}^{TI}_{F}(I_{f})+\mathcal{L}^{CF}_{F}(I_{f}).\label{Eq27}
\end{equation}
For the collaboratively optimized fused image $\tilde{I}_{f}$ obtained by Eq.~\eqref{Eq4}, the corresponding visual fusion loss during the second-stage collaborative optimization is represented as $\mathcal{L}_{F}(\tilde{I}_{f})$.

\subsubsection{Semantic Segmentation Loss}
To supervise semantic segmentation learning, we first employ the conventional cross-entropy loss:
\begin{equation}
	\mathcal{L}^{CE}_{S}(I_{s})=-\sum\limits_{r} \ddot{I}_s(r) \log I_{s}(r),\label{Eq28}
\end{equation}
where $\ddot{I}_s$ refers to the segmentation label, $I_{s}$ denotes the classification probability output by the segmentation decoder, and $r$ indicates the subclass of segmentation categories. In addition, we employ the Lov{\'a}sz-Softmax loss~\cite{berman2018lovasz}, which directly optimizes the Intersection over Union (IoU)-oriented region consistency for semantic prediction:
\begin{equation}
	\mathcal{L}^{LS}_{S}(I_{s})=-\frac{1}{R}\sum\limits_{r} \overline{\triangle J^c}(m(r)),\label{Eq29}
\end{equation}
where $\overline{\triangle J^c}$ is the surrogate of the Jaccard loss. $m(r)$ is the vector of pixel classification errors, defined as:
\begin{equation}
	m(r)=\left\{
	\begin{array}{ll}
		1-I_{s}(r),& \text{if }\ddot{I}_s(r)=1, \\
		I_{s}(r),&  \text{otherwise}.
	\end{array} \right.\label{Eq30}
\end{equation}
The semantic segmentation loss is summarized as:
\begin{equation}
	\mathcal{L}_{S}(I_{s})=\mathcal{L}^{CE}_{S}(I_{s})+\mathcal{L}^{LS}_{S}(I_{s}).\label{Eq31}
\end{equation}
During the second-stage collaborative optimization, the semantic segmentation loss for supervising the refined segmentation result $\tilde{I}_{s}$ is represented as $\mathcal{L}_{S}(\tilde{I}_{s})$.

\begin{figure*}[t]
	\centering
	\includegraphics[width=1\linewidth]{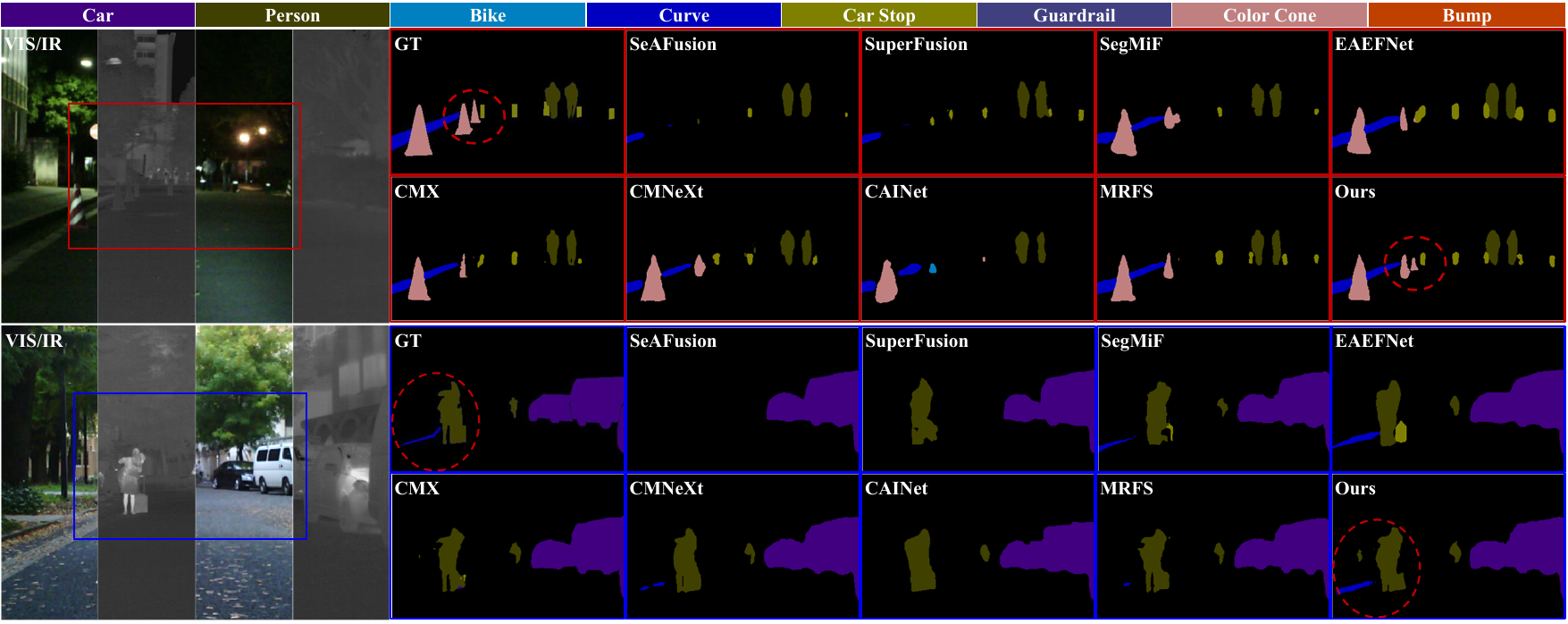}
	\caption{Qualitative results of multi-modal semantic segmentation on the MFNet dataset.}\label{fig10}
	\vspace{-0.05in}
\end{figure*}

\section{Experiments and Results}
In this section, we compare our W2F with state-of-the-art methods on both semantic segmentation and visual fusion tasks. Then, we conduct extensive generalization and application experiments to verify the generalization capability of W2F. At the end, comprehensive ablation studies are conducted to analyze the role of the proposed technical designs.

\subsection{Experimental Configuration}
\subsubsection{Datasets}
We evaluate the performance of our W2F on the MFNet~\cite{ha2017mfnet} and FMB~\cite{liu2023multi} datasets, which contain $1,569$ and $1,500$ spatially aligned visible and infrared image pairs with resolutions of $480\times640$ and $600\times800$, respectively.

\subsubsection{Training Details}
We adopt the lightweight CNN architecture to construct the visual fusion decoder, as it has demonstrated effectiveness in low-level vision tasks~\cite{pan2022dual}. For semantic segmentation, we employ the MLP decoder from SegFormer~\cite{xie2021segformer}. Besides, the extractors in Fig.~\ref{fig2} follow the transformer-based SegFormer architecture, where the number of encoder stages is set to $4$ according to its original configuration. The training of W2F consists of two stages. In the first stage, the visual fusion network and the semantic segmentation network are independently trained using $\mathcal{L}_{F}(I_{f})$ and $\mathcal{L}_{S}(I_{s})$, respectively. The initial learning rate is set to $6e^{-5}$, with a batch size of $4$, and the models are trained for a total of $500$ epochs. In the second stage, the visual fusion encoder and the segmentation encoder are frozen, while the feature search modules are optimized and the corresponding decoders are fine-tuned to achieve visual-semantic collaboration. At this stage, the learning rate is reduced to $5e^{-6}$, and the models are optimized for additional $50$ epochs. All experiments are conducted on four NVIDIA GeForce RTX 3090 GPUs and one AMD EPYC 7H12 64-Core Processor CPU.

\subsection{Semantic Segmentation Comparison}
We first validate our W2F on the multi-modal semantic segmentation task. Eight state-of-the-art competitors are selected, including SeAFusion~\cite{tang2022image}, SuperFusion~\cite{tang2022superfusion}, SegMiF~\cite{liu2023multi}, EAEFNet~\cite{liang2023explicit}, CMX~\cite{zhang2023cmx}, CMNeXt~\cite{zhang2023delivering}, CAINet~\cite{lv2024context}, and MRFS~\cite{zhang2024mrfs}. Among them, SeAFusion, SuperFusion, and SegMiF follow the sequential iterative optimization paradigm, while EAEFNet, CMX, CMNeXt, and CAINet~\cite{lv2024context} adopt the independent optimization paradigm. MRFS is the preliminary version of W2F.

\begin{table}[!t]
	\caption{Quantitative results of multi-modal semantic segmentation on the MFNet dataset.}\label{table-1}
	\vspace{-0.05in}
	\resizebox{\linewidth}{!}{
		\setlength\tabcolsep{1pt}
		\footnotesize
		\centering
		\begin{tabular}{c|cccccccc|c}
			\toprule
			MFNet & Car & Person & Bike & Curve & Car Stop & Guard. & Cone & Bump  & mIoU \\ \midrule
			SeAFusion    & 84.2   & 71.1   & 58.7   & 33.1   & 20.1    & 0.0     & 40.4     & 33.9    & 48.8       \\
			SuperFusion     & 82.9   &69.1    &58.8    &31.3   &27.1     &0.0      &40.0      &37.2     &49.3        \\
			SegMiF        & 87.8   & 71.4   & 63.2   & 47.5   & 31.1    & 0.0     & 48.9     & 50.3    & 56.1       \\ \hline
			EAEFNet       & 87.6   & 72.6   & 63.8   & 48.6   & 35.0    & 14.2    & 52.4     & 58.3    & 58.9       \\
			CMX          &90.1    &75.2    &64.5    &50.2    &35.3     &8.5      &54.2      &60.6     &59.7        \\
			\rowcolor{blue!10}
			CMNeXt           &90.3    &74.5    &64.5    &46.6    &39.2     &15.1      &52.6      &59.8     &\underline{60.1}        \\
			CAINet           &88.5    &66.3    &68.7    &55.4    &31.5     &9.0      &48.9      &60.7     &58.6        \\ \hline
			MRFS          & 89.4   & 75.4   &65.0   & 49.0   & 37.2    & 5.4     & 53.1     & 58.8    &59.1                  \\		
			\rowcolor{pink!40}
			Ours         &89.9   &74.7    &66.5   &47.1    &53.3     &15.8      &54.3      &53.3     &\textbf{61.5}                \\ \bottomrule
	\end{tabular}}
\end{table}

\begin{figure*}[t]
	\centering
	\includegraphics[width=1\linewidth]{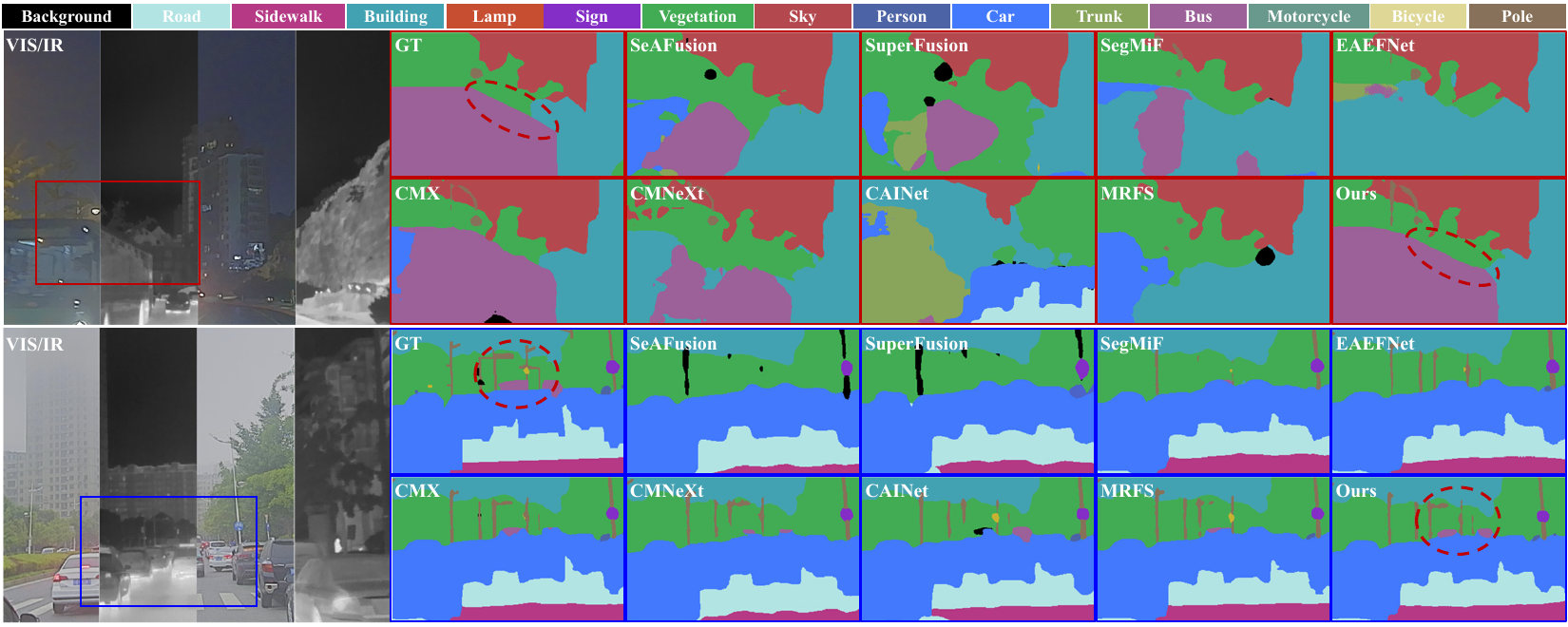}
	\caption{Qualitative results of multi-modal semantic segmentation on the FMB dataset.}\label{fig11}
\end{figure*}

\begin{table}[!t]
	\caption{Quantitative results of multi-modal semantic segmentation on the FMB dataset.}\label{table-2}
	\renewcommand\arraystretch{0.97}
	\vspace{-0.05in}
	\resizebox{\linewidth}{!}{
		\setlength\tabcolsep{1pt}
		\footnotesize
		\centering
		\begin{tabular}{c|cccccccc|c}
			\toprule
			FMB       & Car  &Person  & Truck      & T-Lamp    & T-Sign     & Buil.   & Vege. &Pole &mIoU \\  \midrule
			SeAFusion  & 76.2   & 59.6   & 15.1          & 34.4   & 68.0   & 80.1   & 83.5  & 38.4    & 48.4           \\
			SuperFusion  &75.8    &55.7    &0.06          & 27.3   &69.0   &76.9   &83.2   &36.7     &46.3             \\
			
			SegMiF     & 78.7   & 65.5   & 42.4  & 35.6   & 71.7   & 80.1   & 85.1  & 35.7    & 54.8            \\ \hline

			EAEFNet    & 79.7   & 61.6          & 22.5   & 34.3   & 74.6   & 82.3  & 86.6    & 46.2   & 54.1   \\
			\rowcolor{blue!10}
			CMX  &79.2    &69.1    &39.7          &49.7    &76.5   &84.5   &87.2   &54.1     &\underline{57.6}             \\
			
			CMNeXt  &79.4    &62.3    &32.7          &40.9    &74.4   &81.6   &86.0   &49.4     &56.3             \\
			CAINet  &79.4    &65.5    &30.4          &50.6    &76.1   &81.8   &85.7   &47.8     &54.2             \\
			\hline
			
			MRFS       & 76.2   & 71.3          & 34.4   &50.1   & 75.8   & 85.4                     & 87.0          & 53.6     & 57.1              \\
			\rowcolor{pink!40}
			Ours         &83.1    &72.5    &45.8          &47.1    &77.7   &85.2   &87.9   &54.7     &\textbf{61.7}                  \\ \bottomrule
		\end{tabular}
	}
\end{table}

\subsubsection{Results on the MFNet Dataset}
We first present the qualitative results of multi-modal semantic segmentation on the MFNet dataset in Fig.~\ref{fig10}. W2F demonstrates a relatively fine-grained segmentation performance. For example, in the first set of results, W2F accurately identifies the small-scale color cones and the distant pedestrian, while most competitors fail to detect them. In the second set of results, W2F clearly captures the posture of the person beside the lane line, whereas other methods present confusion. The quantitative results in Table~\ref{table-1} further demonstrate the advantage of W2F. Specifically, W2F achieves the highest mIoU score, outperforming the second-place CMNeXt by $1.4$ percentage points. These results verify the effectiveness of our W2F for multi-modal semantic segmentation on the MFNet dataset.

\begin{figure*}[t]
	\centering
	\includegraphics[width=1\linewidth]{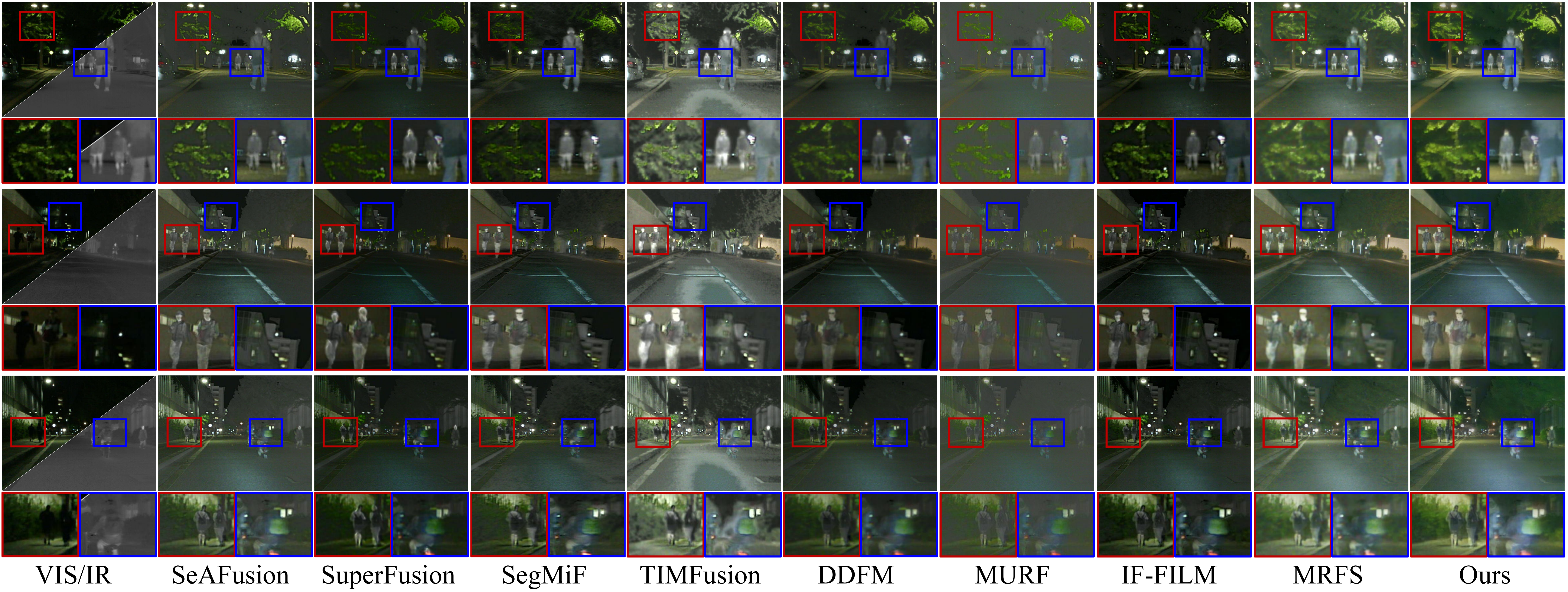}
	\vspace{-0.2in}
	\caption{Qualitative results of multi-modal visual fusion on the MFNet dataset.}\label{fig12}
\end{figure*}

\begin{table}[t]
	\caption{Quantitative results of multi-modal visual fusion on the MFNet dataset.}\label{table-3}
	\resizebox{\linewidth}{!}{
		\setlength\tabcolsep{0.9pt}
		\footnotesize
		\centering
		\begin{tabular}{c|cccc|ccc|cc}
			\toprule
			MFNet & SeAF. & SuperF.  & SegMiF  & TIMF.  & DDFM  & MURF   & IF-F. & MRFS  & Ours \\ \midrule
			VIF $\uparrow$       &\cellcolor{blue!10} \underline{0.720} & 0.531   & 0.565   & 0.702  & 0.718   & 0.434    & 0.719              & 0.716    & \cellcolor{pink!40} \textbf{0.742}       \\
			SCD  $\uparrow$    & 1.057   &0.831  &1.091  & 1.021  &\cellcolor{blue!10} \underline{1.169}  &0.697  &0.981                        &1.068     & \cellcolor{pink!40} \textbf{1.180}        \\
			SSIM $\uparrow$    & \cellcolor{blue!10} \underline{0.746}   & 0.673  & 0.677 & 0.717 & 0.744    & 0.724   & 0.667              & 0.715    & \cellcolor{pink!40} \textbf{0.777}       \\
			$N^{AB/F}$  $\downarrow$   & 0.077   & \cellcolor{blue!10} \underline{0.035} & 0.138   & 0.166 & \cellcolor{pink!40} \textbf{0.023}      & 0.041   & 0.056             & 0.045    & \cellcolor{pink!40} \textbf{0.023}       \\  \bottomrule
		\end{tabular}
	}
\end{table}

\subsubsection{Results on the FMB Dataset}
We further evaluate the semantic segmentation performance on the FMB dataset. The qualitative results are presented in Fig.~\ref{fig11}. In the first set of results, W2F successfully segments the regular contour of the bus, almost perfectly matching the ground truth. In contrast, other methods fail to achieve complete and accurate segmentation in this region. In the second set of results, recognizing small-scale distant objects in hazy weather is challenging. Nevertheless, W2F still accurately identifies distant lamp poles and buses, while other methods suffer from spatial discontinuity and missed detections. The quantitative results are reported in Table~\ref{table-2}. W2F achieves the highest segmentation accuracy, outperforming our preliminary version, MRFS, by $4.6$ percentage points in terms of mIoU. Notably, in MRFS, we excluded the ``bicycle'' class, which does not exist in the testing set, when calculating mIoU. However, in this paper, we also include this class in the statistics. This is why the mIoU of MRFS in Table~\ref{table-2} appears different from that in~\cite{zhang2024mrfs}. However, since the same standard is applied, both are absolutely fair for comparison. Overall, these results demonstrate the effectiveness of W2F for multi-modal semantic segmentation on the FMB dataset.

\begin{figure*}[t]
	\centering
	\includegraphics[width=1\linewidth]{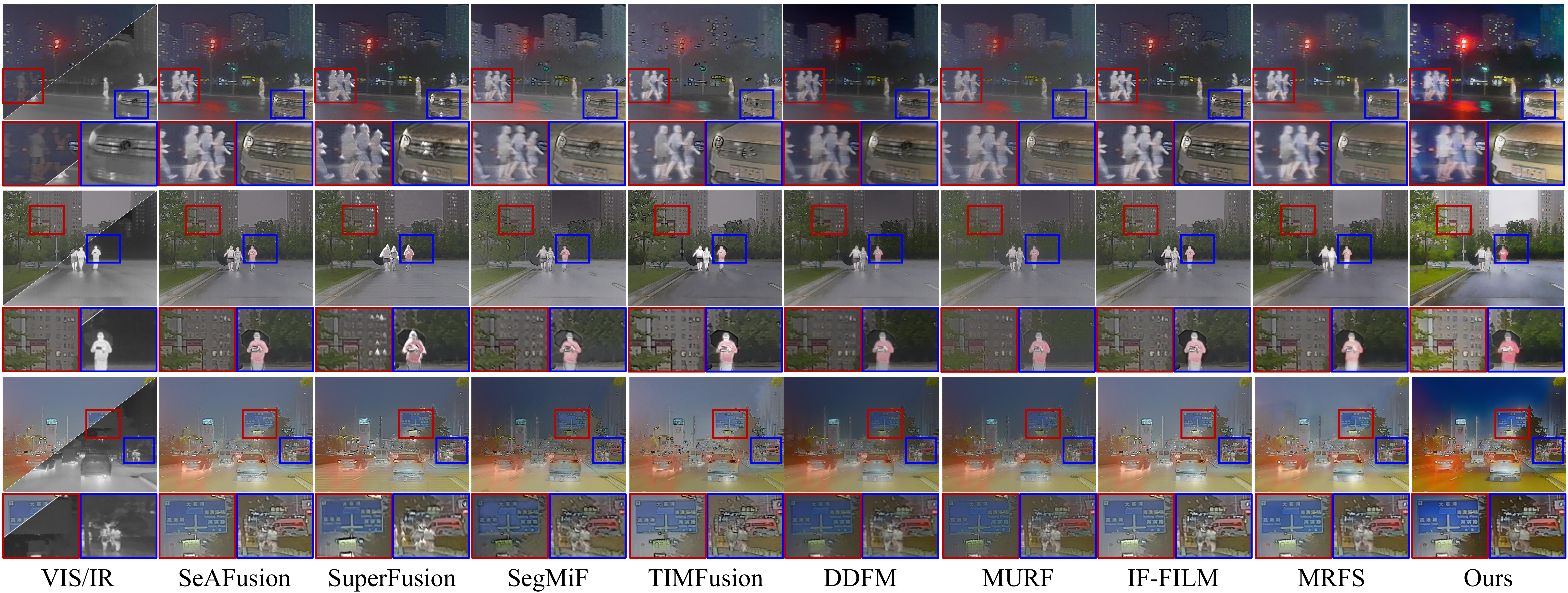}
	\vspace{-0.2in}
	\caption{Qualitative results of multi-modal visual fusion on the FMB dataset.}\label{fig13}
\end{figure*}

\subsection{Visual Fusion Comparison}
We further evaluate W2F on the visual fusion task using both the MFNet and FMB datasets. Eight state-of-the-art visual fusion  methods are selected, including SeAFusion~\cite{tang2022image}, SuperFusion~\cite{tang2022superfusion}, SegMiF~\cite{liu2023multi}, TIMFusion~\cite{liu2024task}, DDFM~\cite{zhao2023ddfm}, MURF~\cite{xu2023murf}, IF-FILM~\cite{zhao2024image}, and MRFS~\cite{zhang2024mrfs}. Among them, SeAFusion, SuperFusion, SegMiF, and TIMFusion follow the sequential iterative optimization paradigm, while DDFM, IF-FILM, and MURF adopt the independent optimization paradigm. We employ four quantitative metrics for objective evaluation, including visual information fidelity (VIF)~\cite{han2013new}, SCD~\cite{aslantas2015new}, structural similarity index measure (SSIM)~\cite{wang2002universal}, and noise-based fusion score ($N^{AB/F}$)~\cite{shreyamsha2013multifocus}.

\begin{table}[t]
	\caption{Quantitative results of multi-modal visual fusion on the FMB dataset.}\label{table-4}
	\resizebox{\linewidth}{!}{
		\setlength\tabcolsep{0.9pt}
		\footnotesize
		\centering
		\begin{tabular}{c|cccc|ccc|cc}
			\toprule
			FMB & SeAF. & SuperF.  & SegMiF  & TIMF.  & DDFM   & MURF  & IF-F.  & MRFS  & Ours \\ \midrule
			VIF $\uparrow$       &0.624   &0.453    &0.556   &0.474   &0.624   &0.545   &\cellcolor{blue!10} \underline{0.639}  &0.632   & \cellcolor{pink!40} \textbf{0.702}        \\
			SCD  $\uparrow$    &0.835   &0.741    &0.824   &0.353   & \cellcolor{blue!10} \underline{0.987}  &0.715  &0.695  &0.704   & \cellcolor{pink!40} \textbf{1.063}        \\
			SSIM $\uparrow$    &0.675   &0.571    &0.704   &0.661   &\cellcolor{blue!10} \underline{0.741}   &0.694   &0.674   &0.690   & \cellcolor{pink!40} \textbf{0.757}       \\
			$N^{AB/F}$  $\downarrow$   &0.094   &0.153    &0.109   &0.101   &\cellcolor{pink!40} \textbf{0.020}   &0.110   &0.096   &0.053   & \cellcolor{blue!10} \underline{0.052}       \\  \bottomrule
		\end{tabular}
	}
\end{table}

\subsubsection{Results on the MFNet Dataset}
The MFNet dataset contains challenging degradation factors such as low light, noise, color distortion, and low contrast. These negative factors severely limit human visual perception and also pose significant challenges to multi-modal visual fusion. In such adverse cases, our W2F still demonstrates strong degradation removal and information fusion capabilities, as shown in Fig.~\ref{fig12}. Specifically, SeAFusion, DDFM, SegMiF, SuperFusion, IF-FILM, and MURF struggle under low-light conditions. By comparison, TIMFusion, MRFS, and W2F can recover scene texture details, such as the tree canopy in the first set of results and the shrubs in the third. Moreover, compared with TIMFusion and MRFS, W2F further produces cleaner and sharper fusion results with fewer artifacts. For example, in the second set of results, W2F better preserves fine details such as the pedestrian’s glasses while avoiding structural blurring. Quantitative results in Table~\ref{table-3} also show that W2F achieves the best scores across all four metrics, demonstrating its effectiveness.

\begin{figure*}[t]
	\centering
	\includegraphics[width=1\linewidth]{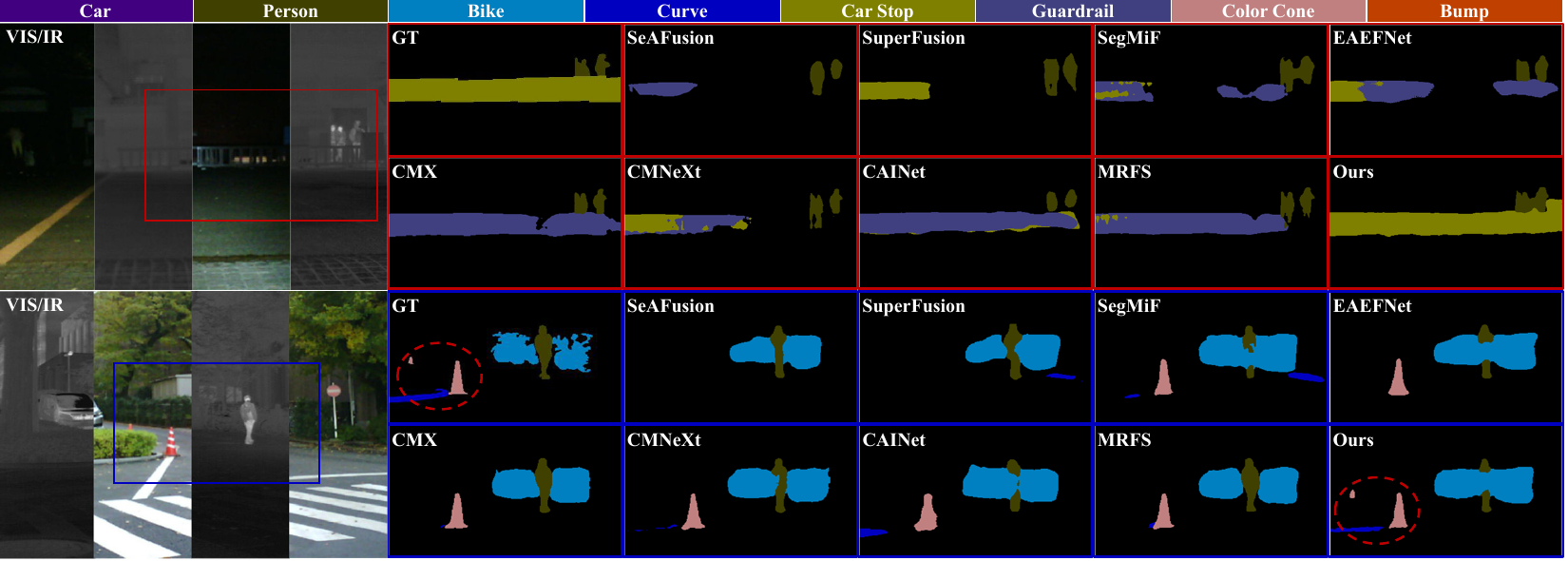}
	\vspace{-0.15in}
	\caption{Qualitative results of semantic segmentation retraining on the MFNet dataset.}\label{fig13-1}
	\vspace{-0.05in}
\end{figure*}

\subsubsection{Results on the FMB Dataset}
The FMB dataset is mainly affected by low light, color distortion, and haze, resulting in poor scene visibility. Under such challenging conditions, W2F still shows strong visual robustness, as shown in Fig.~\ref{fig13}. First, W2F effectively enhances low-light scenes while seamlessly integrating thermal radiation objects into the environment. For example, in the first set of results, the details of the car are effectively illuminated and the license plate becomes recognizable. Meanwhile, thermal pedestrians are distinctly highlighted on the road. Second, W2F effectively eliminates the hazy appearance and significantly improves scene visibility. In the second set of results, W2F restores clearer trees and buildings in the background, whereas other methods remain noticeably affected by haze. This also makes the foreground thermal pedestrians more salient. Similar observations can be found in the third set of results. The quantitative results in Table~\ref{table-4} further verify the superiority of W2F on the FMB dataset.

\subsection{Retraining Analysis with Restoration Prior}
We conduct additional retraining experiments on comparative methods using the constructed clean datasets to embed information restoration prior for them. Specifically, for semantic segmentation methods, the restored clean images are directly used as network inputs for retraining. For visual fusion methods, the restored clean infrared-visible image pairs are introduced as supervision during training.

\begin{figure*}[t]
	\centering
	\includegraphics[width=1\linewidth]{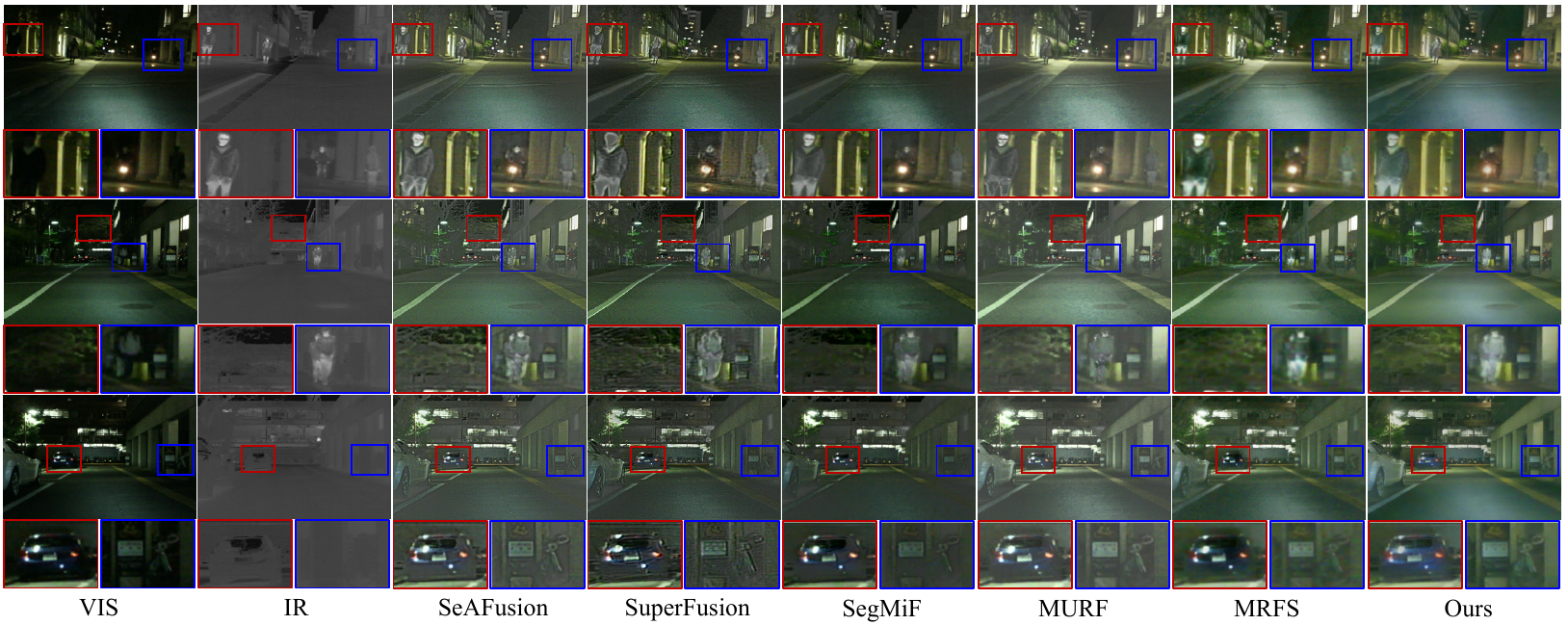}
	\vspace{-0.2in}
	\caption{Qualitative results of visual fusion retraining on the MFNet dataset.}\label{fig13-2}
	\vspace{-0.05in}
\end{figure*}

\subsection{Semantic Segmentation Retraining}
First, we directly use Photoshop-processed clean data as input to train all comparative semantic segmentation methods. This setting can be understood as directly injecting the information restoration prior into the segmentation model at the data level. The experimental results are shown in Fig.~\ref{fig13-1} and Table~\ref{table4-1}. 

\begin{table}[!t]
	\caption{Quantitative results of semantic segmentation retraining on the MFNet dataset.}\label{table4-1}
	\resizebox{\linewidth}{!}{
		\setlength\tabcolsep{1pt}
		\footnotesize
		\centering
		\begin{tabular}{c|cccccccc|c}
			\toprule
			MFNet & Car & Person & Bike & Curve & Car Stop & Guard. & Cone & Bump  & mIoU \\ \midrule
			SeAFusion    & 81.9 & 66.4 & 58.1 & 32.4 & 11.4 & 0.0 & 44.0 & 48.4 & 48.9 \\
			SuperFusion  & 83.1 & 69.4 & 58.2 & 32.2 & 28.4 & 0.0 & 40.2 & 42.3 & 50.2 \\
			SegMiF       & 87.2 & 71.0 & 61.7 & 39.7 & 27.5 & 9.8 & 56.8 & 47.6 & 55.5 \\ \hline
			EAEFNet      & 86.8 & 71.2 & 64.8 & 49.6 & 36.5 & 11.4 & 52.6 & 53.7 & 58.3 \\
			CMX          & 90.0 & 74.8 & 64.8 & 48.3 & 35.3 & 8.9 & 54.4 & 59.5 & 59.4 \\
			\rowcolor{blue!10}
			CMNeXt       & 90.3 & 73.7 & 64.0 & 44.1 & 39.8 & 17.0 & 53.3 & 56.4 & 59.6 \\
			CAINet       & 89.2 & 69.6 & 69.4 & 53.3 & 35.9 & 9.0 & 48.4 & 58.4 & 59.1 \\ \hline
			MRFS         & 89.4 & 75.5 & 65.0 & 49.1 & 37.3 & 5.3 & 53.1 & 58.8 & 59.1 \\
			\rowcolor{pink!40}
			Ours         & 89.9 & 74.7 & 66.5 & 47.1 & 53.3 & 15.8 & 54.3 & 53.3 & \textbf{61.5}
			\\ \bottomrule
		\end{tabular}
	}
	\vspace{-0.05in}
\end{table}

\begin{table}[t]
	\caption{Quantitative results of visual fusion retraining on the MFNet dataset.}\label{table4-2}
	\scalebox{0.82}{
		\setlength\tabcolsep{5pt}
		\footnotesize
		\centering
		\begin{tabular}{c|cccccc}
			\toprule
			MFNet & SeAF. & SuperF. & SegMiF & MURF & MRFS & Ours \\ \midrule
			VIF $\uparrow$
			& 0.676 & 0.342   & 0.663   & 0.716   & \cellcolor{pink!40} \textbf{0.767}    & \cellcolor{blue!10} \underline{0.742}       \\
			SCD $\uparrow$    & 1.010   & 0.814  & 1.004  & 1.126  & \cellcolor{blue!10} \underline{1.160}     & \cellcolor{pink!40} \textbf{1.180}        \\
			SSIM $\uparrow$    & \cellcolor{blue!10} \underline{0.737}   & 0.571  & 0.727 & 0.748    & \cellcolor{blue!10} \underline{0.737}    & \cellcolor{pink!40} \textbf{0.777}       \\
			$N^{AB/F}$ $\downarrow$   & 0.073   & 0.392 & 0.084   &\cellcolor{blue!10} \underline{ 0.070}      & 0.073    & \cellcolor{pink!40} \textbf{0.023}       \\
			\bottomrule
	\end{tabular}}
	\vspace{-0.05in}
\end{table}

It can be observed that directly using the processed clean data does not consistently improve the segmentation performance, and in some cases even leads to comparable or lower results than those obtained using the original degraded inputs. We attribute this phenomenon to the fact that Photoshop-based degradation removal is an offline image-level preprocessing operation. Although it can improve the visual appearance of degraded observations, it may also remove or distort certain modality-specific details and boundary cues that are important for semantic segmentation, thereby limiting or even weakening the segmentation performance. In contrast, our W2F still takes the originally captured degraded multi-modal images as input, thus preserving the modality-specific details and semantic cues contained in the original observations. The restoration prior is learned through clean supervision and then transferred to the semantic fusion branch via the feature search module, providing a more adaptive and less destructive way to benefit semantic segmentation.

\begin{figure*}[t]
	\centering
	\includegraphics[width=1\linewidth]{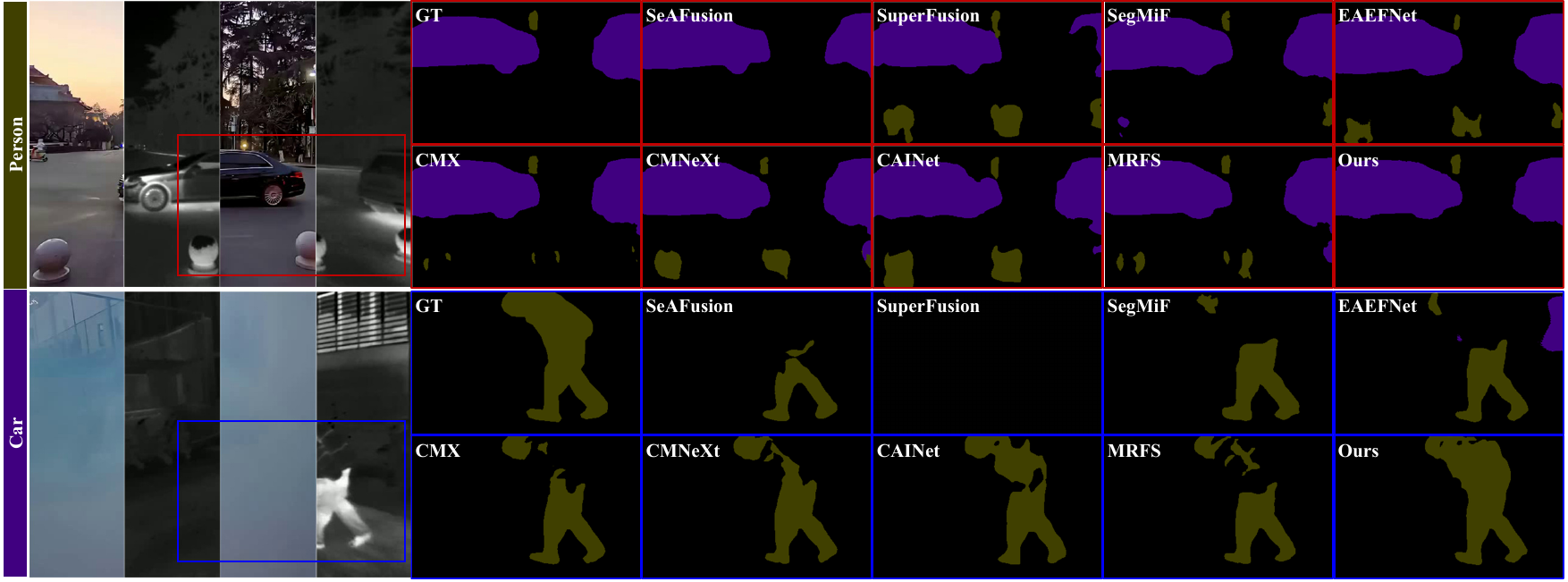}
	\vspace{-0.2in}
	\caption{Qualitative results of generalization to the M3SVD dataset on semantic segmentation task.}\label{fig14-1}
\end{figure*}

\begin{figure*}[t]
	\centering
	\includegraphics[width=1\linewidth]{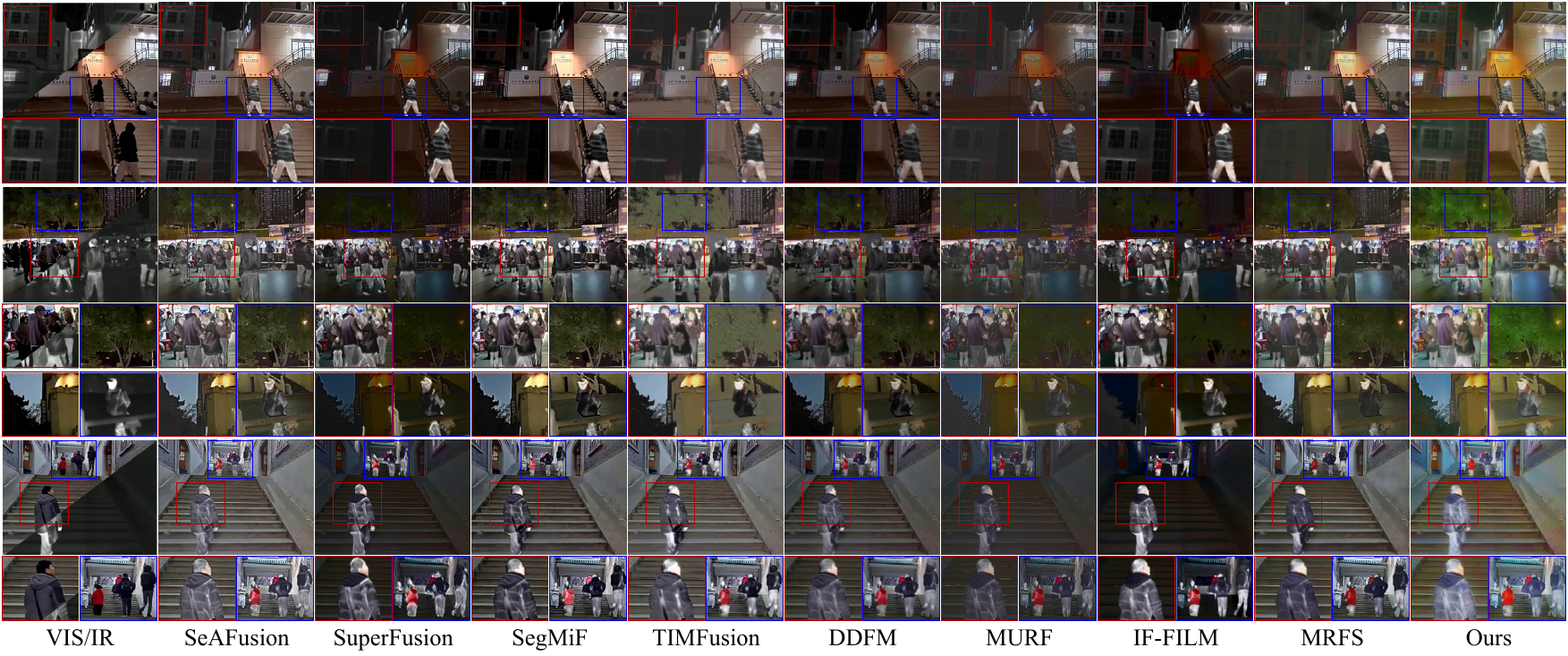}
	\vspace{-0.2in}
	\caption{Qualitative results of generalization to the M3SVD dataset on visual fusion task.}\label{fig14-2}
\end{figure*}

\subsection{Visual Fusion Retraining}
For the visual fusion methods, we further conduct retraining experiments using the constructed clean infrared-visible image pairs as restoration supervision. Specifically, SeAFusion, SuperFusion, SegMiF, TIMFusion, and MRFS are retrained in this experiment. DDFM, MURF, and IF-FILM are excluded because their training code is not publicly available or their training process relies on additional unavailable training data, making retraining infeasible under our setting. The qualitative and quantitative results are shown in Fig.~\ref{fig13-2} and Table~\ref{table4-2}, respectively. 

\begin{table}[!t]
	\caption{Quantitative results of generalization to the M3SVD dataset on multi-modal semantic segmentation task.}\label{table-5-0}
	\vspace{-0.05in}
	\resizebox{\linewidth}{!}{
		\setlength\tabcolsep{1pt}
		\footnotesize
		\centering
		\begin{tabular}{c|ccc|cccc|cc}
			\toprule
			M3SVD       & SeAF.  &SuperF.  & SegMiF      & EAEF.    & CMX     & CMNeXt   & CAINet &MRFS &Ours \\  \midrule
			Car  & 21.5   & 22.2   & 39.3          & 16.5   & 70.8   & \cellcolor{blue!10} 75.4   & 53.2  & 69.1    & \cellcolor{pink!40} 73.7           \\
			Person  &65.4    &61.3    &73.3          & 78.0   &79.5   &\cellcolor{blue!10}77.5   &72.4   &77.1     &\cellcolor{pink!40} 82.3             \\
			\midrule
			mIoU  &43.5    &41.8    &56.3          & 47.3   &75.2   &\cellcolor{blue!10} \underline{76.5}   &62.8   &73.1     &\cellcolor{pink!40} \textbf{77.8}             \\ \bottomrule
		\end{tabular}
	}
\end{table}

As observed, introducing restoration supervision improves the degradation removal capability of existing methods to some extent, particularly in low-light enhancement and scene visibility restoration. However, most retrained methods still suffer from artifacts, structural blurring, and insufficient emphasis on key objects. These results indicate that restoration supervision alone is insufficient for achieving comprehensive visual perception enhancement. Although the introduced restoration prior improves visual restoration capability, it cannot effectively enhance semantically important regions and foreground structures. In contrast, W2F consistently produces cleaner visual appearances and more natural foreground-background integration through the proposed visual-semantic collaboration framework.

\subsection{Generalization Evaluation}
Given that semantic segmentation annotations are unavailable for many infrared-visible datasets, a strong generalization capability is essential for practical deployment. Therefore, we further evaluate whether W2F can maintain effective visual fusion and semantic perception performance under unseen scenarios without retraining.

\begin{figure*}[t]
	\centering
	\includegraphics[width=1\linewidth]{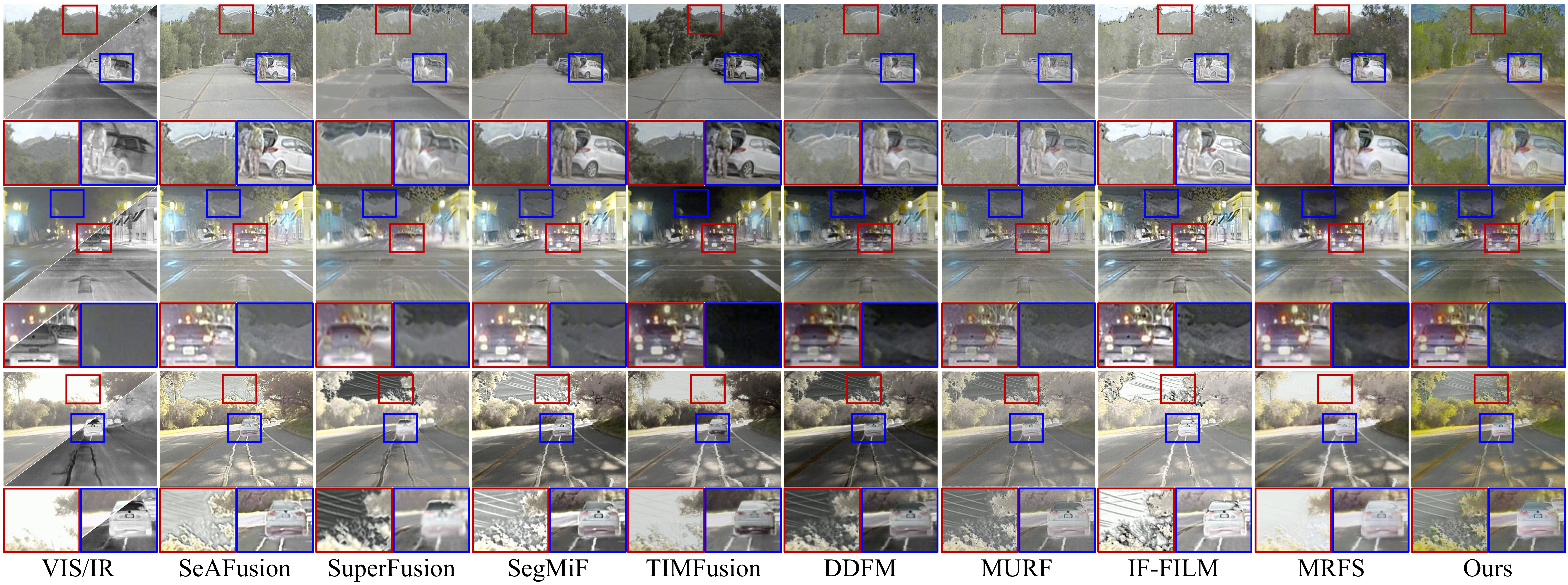}
	\vspace{-0.2in}
	\caption{Qualitative results of generalization to the RoadScene dataset.}\label{fig14-3}
	\vspace{-0.1in}
\end{figure*}

\begin{table}[!t]
	\caption{Quantitative results of generalization to the M3SVD dataset on visual fusion task.}\label{table-5-1}
	\vspace{-0.05in}
	\resizebox{\linewidth}{!}{
		\setlength\tabcolsep{0.9pt}
		\footnotesize
		\centering
		\begin{tabular}{c|cccc|ccc|cc}
			\toprule
			RoadScene & SeAF. & SuperF.  & SegMiF  & TIMF.  & DDFM  & MURF   & IF-F. & MRFS  & Ours \\ \midrule
			VIF $\uparrow$      &\cellcolor{pink!40} \underline{0.978}    &0.601    &\cellcolor{blue!10} \textbf{1.016}     &0.635    & 0.881    &0.506   &0.916     &0.915     &0.799       \\
			SCD  $\uparrow$    &1.352    &1.472    &1.330     &1.318    &\cellcolor{blue!10} \underline{1.597}    &1.245   &0.801     &1.238     & \cellcolor{pink!40} \textbf{1.648}       \\
			SSIM $\uparrow$    &0.599    &0.579    &0.599     &0.600    &\cellcolor{blue!10} \underline{0.643}    &0.584   &0.613     &0.639     &\cellcolor{pink!40} \textbf{0.656}      \\
			$N^{AB/F}$  $\downarrow$       &0.352    &0.109    &0.399     &0.209    &0.106    &0.149   &\cellcolor{blue!10} \textbf{0.047}     &0.238     &\cellcolor{pink!40} \underline{0.104}       \\  \bottomrule
		\end{tabular}
	}
	\vspace{-0.1in}
\end{table}

\subsubsection{Generalization on M3SVD dataset}
To verify the generalization performance of our W2F, we directly test \textbf{the model trained on the MFNet dataset} on the unseen M3SVD dataset~\cite{Tang2026VideoFusion}. The qualitative segmentation results are presented in Fig.~\ref{fig14-1}. Our W2F demonstrates advanced performance by accurately identifying cars and pedestrians in the scenes, achieving the highest consistency with segmentation labels. Besides, the quantitative metrics in Table~\ref{table-5-0} further corroborate the exceptional generalization capability of our method. In the visual fusion task, W2F consistently produces visually compelling fusion outputs, particularly excelling in color fidelity, contrast, and texture richness, as shown in Fig.~\ref{fig14-2}. Specifically, when details in the source visible images are obscured under nighttime conditions, our W2F successfully restores scene tones and intricate textures, such as the tree canopies and building windows hidden in darkness. In the third set of results, W2F achieves the most balanced scene color tones and provides the clearest and most detailed representations of persons, with enhanced contrast and definition. The quantitative metrics in Table~\ref{table-5-1} further substantiate the excellent visual fusion performance. Overall, the above results on the M3SVD dataset underscore the robust generalization capability of W2F across unseen multi-modal perception scenarios.

\begin{table}[t]
	\caption{Quantitative results of generalization to the RoadScene dataset.}\label{table-5-2}
	\vspace{-0.05in}
	\resizebox{\linewidth}{!}{
		\setlength\tabcolsep{0.9pt}
		\footnotesize
		\centering
		\begin{tabular}{c|cccc|ccc|cc}
			\toprule
			M3SVD & SeAF. & SuperF.  & SegMiF  & TIMF.  & DDFM  & MURF   & IF-F. & MRFS  & Ours \\ \midrule
			VIF $\uparrow$      &0.692    &0.668    &\cellcolor{blue!10} \underline{0.801}     &0.785    & \cellcolor{pink!40} \textbf{0.822}    &0.769   &0.760     &0.751     &0.690       \\
			SCD  $\uparrow$    &1.487    &1.015    &1.702     &1.070    &\cellcolor{blue!10} \underline{1.714}    &1.650   &1.298     &1.267     & \cellcolor{pink!40} \textbf{1.726}       \\
			SSIM $\uparrow$    &0.614    &0.642    &0.659     &0.685    &\cellcolor{blue!10} \underline{0.708}    &0.696   &0.609     &0.679     &\cellcolor{pink!40} \textbf{0.748}      \\
			$N^{AB/F}$  $\downarrow$       &0.386    &0.292    &0.368     &0.257    &\cellcolor{blue!10} \underline{0.242}    &0.304   &0.383     &0.252     &\cellcolor{pink!40} \textbf{0.209}       \\  \bottomrule
		\end{tabular}
	}
\end{table}

\subsubsection{Generalization on RoadScene dataset}
Similarly, we further directly test \textbf{the model trained on the MFNet dataset} on the unseen RoadScene dataset~\cite{xu2022u2fusion} without any additional retraining or fine-tuning. Since the RoadScene dataset does not provide semantic segmentation annotations, we only conduct comparisons on the visual fusion task. The qualitative results are presented in Fig.~\ref{fig14-3}.

\begin{figure*}[t]
	\centering
	\includegraphics[width=1\linewidth]{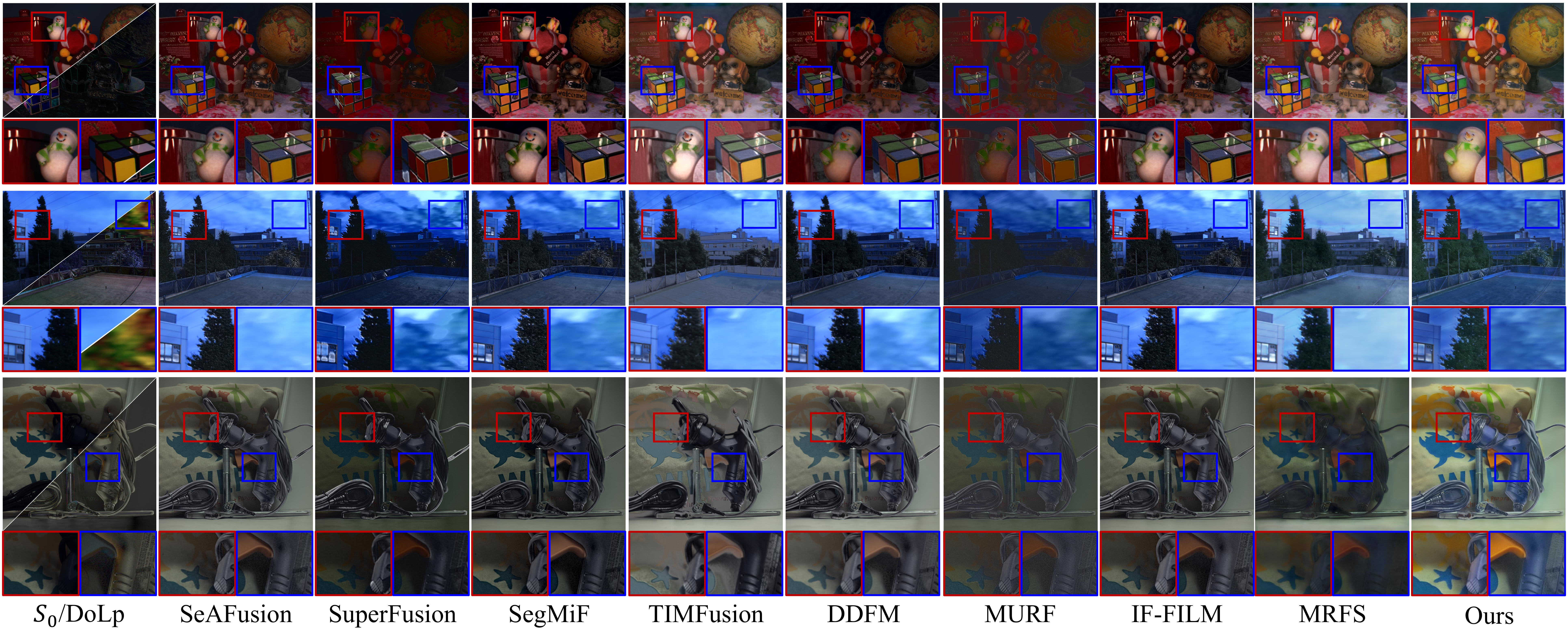}
	\vspace{-0.2in}
	\caption{Qualitative results of application to the polarization image fusion task.}\label{fig15-1}
	\vspace{-0.1in}
\end{figure*}

\begin{figure*}[t]
	\centering
	\includegraphics[width=1\linewidth]{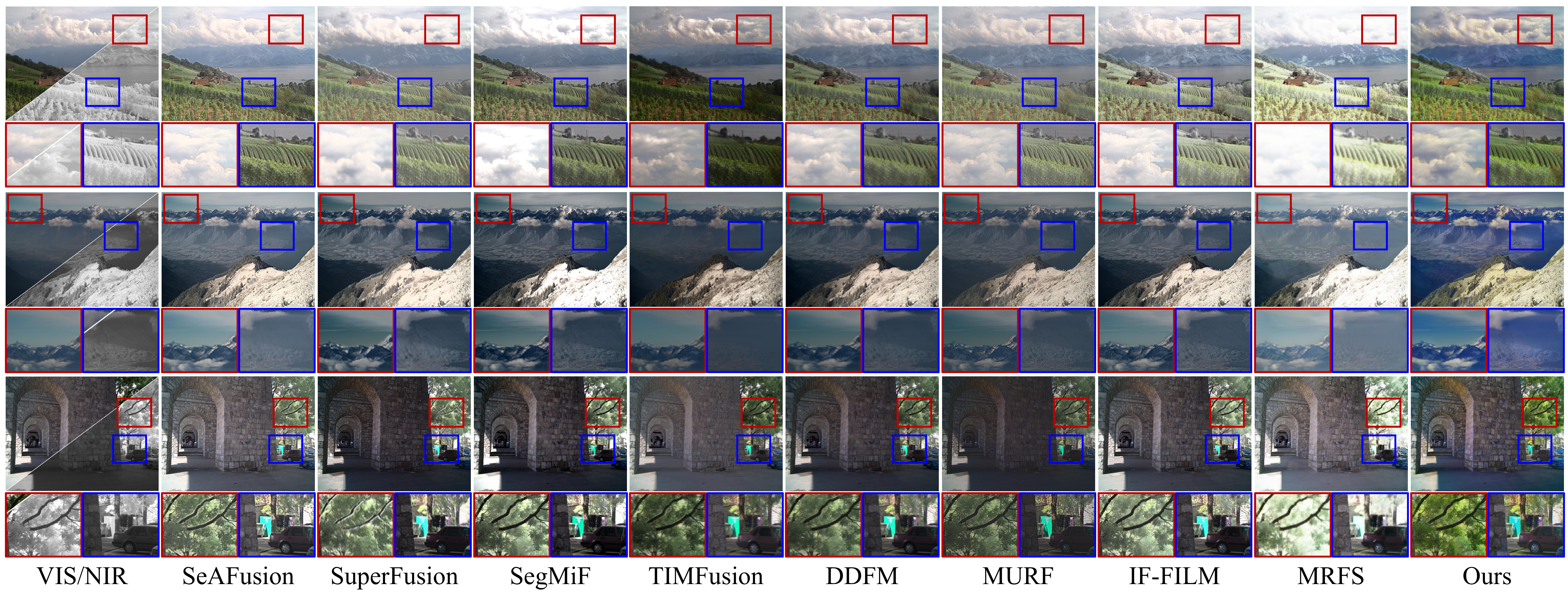}
	\vspace{-0.2in}
	\caption{Qualitative results of application to the near-infrared and visible image fusion task.}\label{fig15-2}
\end{figure*}

\begin{table}[t]
	\caption{Quantitative results of application to the polarization visual fusion task.}\label{table-5-3}
	\resizebox{\linewidth}{!}{
		\setlength\tabcolsep{0.9pt}
		\footnotesize
		\centering
		\begin{tabular}{c|cccc|ccc|cc}
			\toprule
			Polarization & SeAF. & SuperF.  & SegMiF  & TIMF.  & DDFM  & MURF   & IF-F. & MRFS  & Ours \\ \midrule
			VIF $\uparrow$       &0.860    &0.451    &\cellcolor{pink!40} \textbf{0.951}     &0.800     &0.848      & 0.412     &\cellcolor{blue!10} \underline{0.937}      &0.833      &0.772       \\
			SCD  $\uparrow$    &1.417    &1.170    &1.392     &1.098     &\cellcolor{blue!10} \underline{1.576}      &1.043      &1.343      &1.132      &\cellcolor{pink!40} \textbf{1.585}       \\
			SSIM $\uparrow$    &\cellcolor{blue!10} \underline{0.660}    &0.606    &0.635     &0.599     &0.656      &0.624      &0.659      &0.619      &\cellcolor{pink!40} \textbf{0.661}      \\
			$N^{AB/F}$  $\downarrow$   &0.149    &\cellcolor{blue!10} \underline{0.103}    &0.207     &0.252     &0.109      &0.254      &0.180      &0.126      &\cellcolor{pink!40} \textbf{0.095}    \\  \bottomrule
		\end{tabular}
	}
\end{table}

It can be observed that our W2F produces visually appealing fusion results with superior color fidelity, contrast preservation, and texture representation. In the first and third sets of results, the source visible images suffer from reduced contrast caused by slight overexposure, leading most comparative methods to generate fused images with dull and low-saturation scene appearances. By comparison, W2F restores more natural and delicate scene tones while preserving clearer structural details, such as the distant tree canopies and power lines. In the second set of results, severe nighttime light pollution introduces obvious haze-like degradation into the road scene, significantly affecting scene visibility. Even under this challenging condition, W2F still maintains high visual contrast and clear structural representation in the fused image, sharply preserving important details of the sedan and the distant mountains. Moreover, thermal objects remain naturally integrated with the surrounding environment without introducing noticeable artifacts or structural inconsistencies. Furthermore, the quantitative results in Table~\ref{table-5-2} also confirm the high generalization performance of our method.

\subsection{Application to Other Fusion Tasks}
Under the constraints of the segmentation-guided fusion loss and information restoration prior, our W2F learns strong scene information integration and degradation removal capabilities. These abilities are broadly beneficial across different image fusion tasks. To further evaluate the cross-task transferability of W2F, we directly transfer \textbf{the model trained on the infrared-visible MFNet dataset} to three other fusion tasks, including polarization image fusion, near-infrared and visible image fusion, and medical image fusion, without any additional retraining or fine-tuning. In other words, all experiments in this section are conducted under fixed model parameters. Through these experiments, we aim to investigate whether W2F can still maintain effective fusion performance under unseen fusion tasks beyond infrared-visible perception.

\begin{figure*}[t]
	\centering
	\includegraphics[width=1\linewidth]{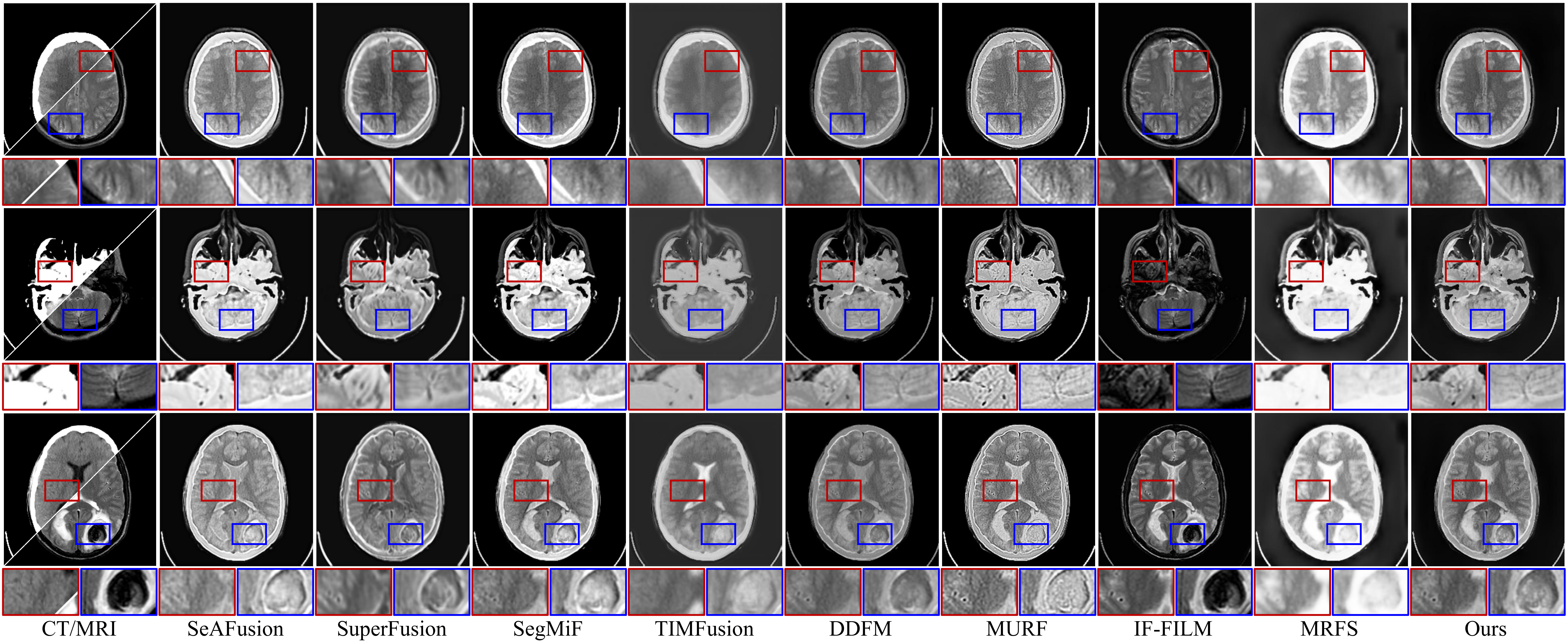}
	\vspace{-0.2in}
	\caption{Qualitative results of application to the medical image fusion task.}\label{fig15-3}
\end{figure*}

\begin{table}[t]
	\caption{Quantitative results of application to the near-infrared and visible image fusion task.}\label{table-5-4}
	\resizebox{\linewidth}{!}{
		\setlength\tabcolsep{0.9pt}
		\footnotesize
		\centering
		\begin{tabular}{c|cccc|ccc|cc}
			\toprule
			NIR-VIS & SeAF. & SuperF.  & SegMiF  & TIMF.  & DDFM  & MURF   & IF-F. & MRFS  & Ours \\ \midrule
			VIF $\uparrow$       &1.246    &0.940    &1.318     &1.047     &\cellcolor{blue!10} \underline{1.319}  &0.994      &1.315      &1.162      &\cellcolor{pink!40} \textbf{1.331}        \\
			SCD  $\uparrow$    &1.181    &1.000    & \cellcolor{pink!40} \textbf{1.522}     &0.780     & \cellcolor{blue!10} \underline{1.310}     &0.562      &1.303      &1.184      &0.596        \\
			SSIM $\uparrow$    &0.801    &0.738    &0.779     &0.781     &\cellcolor{blue!10} \underline{0.830}      &0.788      &0.820      &0.775      &\cellcolor{pink!40} \textbf{0.878}      \\
			$N^{AB/F}$  $\downarrow$  &0.186    &0.149    &0.185     &0.100     &\cellcolor{pink!40} \textbf{0.053} &0.181      &0.166      &0.128      &\cellcolor{blue!10} \underline{0.091}      \\  \bottomrule
		\end{tabular}
	}
\end{table}

\subsubsection{Polarization Image Fusion}
Polarization image fusion refers to the technique of combining the intensity images $S_0$ with the degree of linear polarization (DoLP) images, to overcome the limitations of traditional imaging systems in perceiving surface features of the objects. We attempt to use W2F to address the polarization image fusion problem and implement comparisons on public datasets~\cite{liu2023polarization,morimatsu2020monochrome}. The qualitative results are shown in Fig.~\ref{fig15-1}, in which W2F demonstrates visual performance superior to other competitors. First, our proposed method achieves high-quality color fidelity, as seen in the Rubik's cube in the first set of results and the handbag in the third set of results. Second, W2F achieves the most refined and complete integration of texture details, such as the clouds in the second set of results. More importantly, W2F effectively removes low-light degradation, enhancing scene visibility. The quantitative results in Table~\ref{table-5-3} further validate the advantages of W2F. In general, these results demonstrate that W2F can effectively solve the polarization image fusion task.

\begin{table}[t]
	\caption{Quantitative results of application to the medical image fusion task.}\label{table-5-5}
	\resizebox{\linewidth}{!}{
		\setlength\tabcolsep{0.9pt}
		\footnotesize
		\centering
		\begin{tabular}{c|cccc|ccc|cc}
			\toprule
			Medical & SeAF. & SuperF.  & SegMiF  & TIMF.  & DDFM  & MURF   & IF-F. & MRFS  & Ours \\ \midrule
			VIF $\uparrow$       &0.619    &0.448    &0.635    &0.568     &0.664    &\cellcolor{blue!10} \underline{0.700}     &0.611      &0.598      &\cellcolor{pink!40} \textbf{0.788}               \\
			SCD  $\uparrow$    &0.794    &0.238    &0.695    &0.740     &0.219    &0.735     &\cellcolor{blue!10} \underline{0.983}      & \cellcolor{pink!40} \textbf{1.006}      &0.515          \\
			SSIM $\uparrow$    &0.304    &0.236    &0.723    &0.250     &\cellcolor{blue!10} \underline{0.753}    &0.718     &0.444      &0.268      &\cellcolor{pink!40} \textbf{0.770}          \\
			$N^{AB/F}$  $\downarrow$  &0.234    &0.234    &0.110    &0.366     &\cellcolor{pink!40} \textbf{0.048}    &0.184     &0.231      &0.484      &\cellcolor{blue!10} \underline{0.077}         \\  \bottomrule
		\end{tabular}
	}
\end{table}

\subsubsection{Near-Infrared and Visible Image Fusion}
Unlike our mainly focused infrared and visible image fusion task, the goal of near-infrared (NIR) and visible image fusion is primarily to use the texture structures in the near-infrared image to assist in improving the quality of the visible image. This requires that the fused image enhance the comprehensive representation of the scene as much as possible, without deviating from the visible modality. We provide the qualitative results in Fig.~\ref{fig15-2}. First, W2F presents very clear texture appearances, such as the clouds in the first set of results and the mountains in the second set. By comparison, other comparative methods exhibit localized blurring or even loss of details. Second, W2F is highly faithful to the true color distribution of the objects, such as the bushes in the first set of results and the canopy in the third set. In contrast, other competitors, when incorporating texture details from the near-infrared image, more or less cause the color to appear whitened. We further report the quantitative results in Table~\ref{table-5-4}, where W2F achieves competitive scores. These results demonstrate the excellent handling capability of W2F for the near-infrared and visible image fusion task.

\subsubsection{Medical Image Fusion}
The goal of medical image fusion is to provide a more comprehensive and clearer depiction of the internal structures of the body. We apply W2F to the task of CT and MRI brain image fusion to examine whether it can achieve a comprehensive representation of brain tissues. The qualitative comparison on the Harvard medical dataset\footnote{\url{http://www.med.harvard.edu/AANLIB/home.html}} is presented in Fig.~\ref{fig15-3}. Our W2F effectively filters out noise in CT images while outperforming competitors in brain structure preservation and contrast retention. Besides, The quantitative results in Table~\ref{table-5-5} show that W2F achieves the best scores on most metrics. Overall, W2F can be adapted to address the medical image fusion task well.

\begin{figure}[t]
	\centering
	\includegraphics[width=1\linewidth]{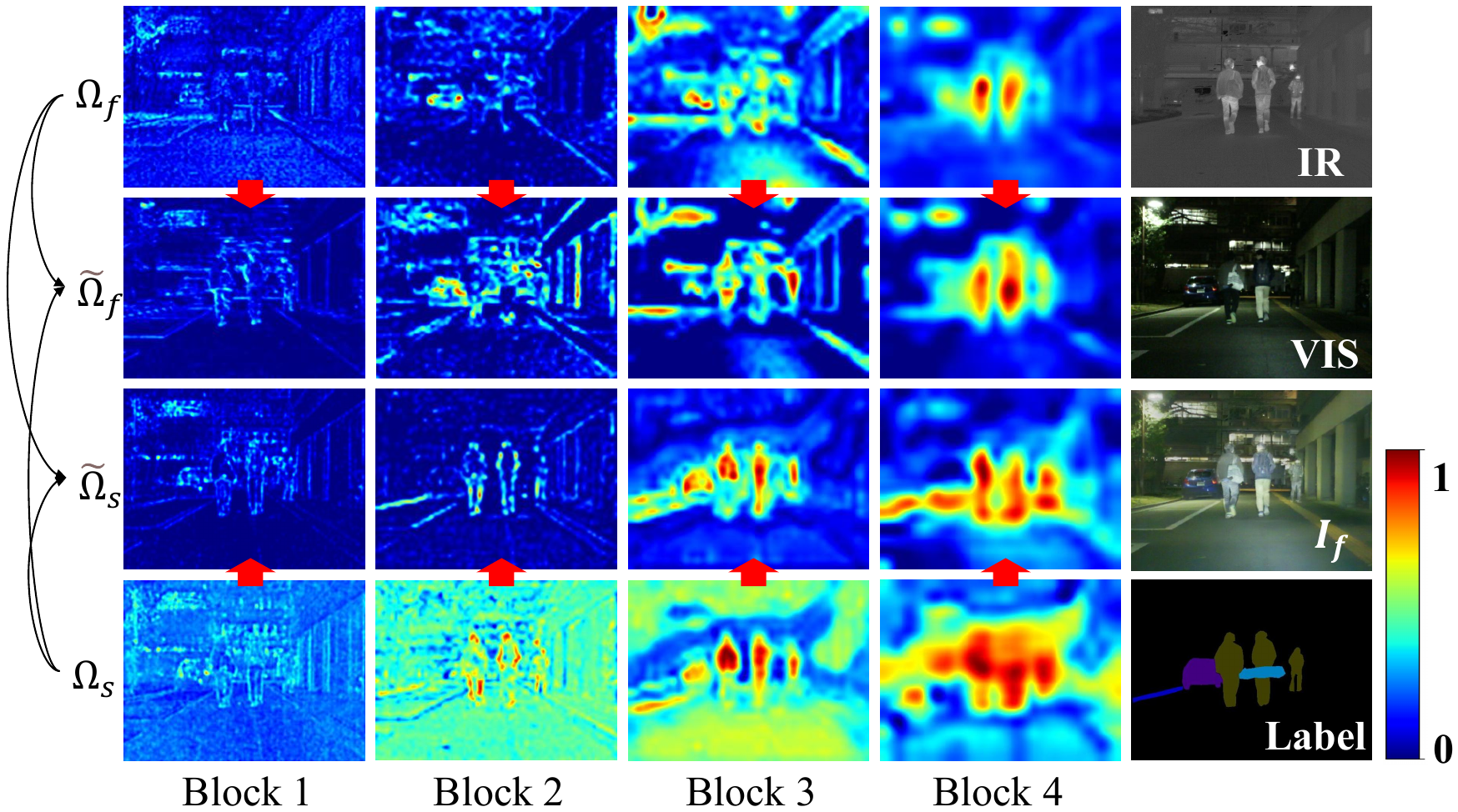}
	\vspace{-0.15in}
	\caption{Feature visualization of visual fusion and semantic fusion before and after interaction. }\label{fig16-0}
\end{figure}

\subsection{Visual-Semantic Collaboration Analysis}
We analyze visual-semantic collaboration from the perspective of feature representations. Specifically, the mutual reinforcement between visual fusion and semantic segmentation in W2F is rooted in two key characteristics: feature differentiation and feature complementarity. Feature differentiation arises because visual fusion primarily captures texture details, intensity contrast, and globally coherent scene structures, whereas semantic fusion emphasizes object-aware structural cues and category-discriminative representations. Feature complementarity follows naturally: the distinct visual and semantic features provide mutually beneficial information, enabling the visual branch to aid semantic reasoning and the semantic branch to guide visual enhancement. Together, these two characteristics form a solid foundation for bidirectional collaboration.

To provide intuitive evidence, we visualize intermediate features at different stages of W2F. As shown in Fig.~\ref{fig16-0}, features (\textit{i.e.}, $\Omega_f$ and $\Omega_s$) from each encoder block of both visual and semantic branches are inspected, along with the updated features (\textit{i.e.}, $\tilde{\Omega}_f$ and $\tilde{\Omega}_s$) after interaction through the feature search module. In the original visual features $\Omega_f$, responses are mainly concentrated on low-level textures, intensity contrast, and global scene structures, with relatively weak and scattered attention to foreground objects, especially in early blocks. Conversely, the original semantic features $\Omega_s$ already highlight object regions and coarse structural layouts but still contain background interference and unstable activations under degraded conditions. After feature interaction, the updated visual features $\tilde{\Omega}_f$ show more focused responses around pedestrians and clearer object contours, indicating that semantic cues guide the visual branch to better emphasize foreground objects and structural details. Meanwhile, updated semantic features $\tilde{\Omega}_s$ become cleaner and more discriminative, with reduced background noise and more complete object-level activations, especially in middle and deep blocks. This demonstrates that the visual features provide the restored and more comprehensive scene representations, allowing the segmentation branch to extract category-relevant cues from less corrupted visual evidence. Overall, these observations confirm that feature differentiation and feature complementarity jointly enable the visual and semantic branches to progressively reinforce each other, supporting effective visual-semantic collaboration throughout the network hierarchy.

\subsection{Ablation Studies}
In this section, we conduct comprehensive ablation studies on the key designs in W2F to verify their contributions, involving \textbf{fourteen} model variants.
\begin{enumerate}
	\item \textbf{Model I}: \textit{Backbone}. In this model, visual fusion and semantic segmentation are independent of each other, with a conventional pooling-based attention module~\cite{deng2021feanet} to achieve a local mutual complement of multi-modal features, and a cross-attention module~\cite{zhou2022multispectral} to achieve a global multi-modal feature aggregation.
	\item \textbf{Model II}: \textit{Backbone $+$ IGM-Att}. This model, based on the backbone, uses the proposed IGM-Att to replace the conventional pooling-based attention module to achieve a local mutual complement of multi-modal features.
	\item \textbf{Model III}: \textit{Backbone $+$ IGM-Att $+$ PC-Att}. Based on Model II, this model further introduces PC-Att to replace the cross-attention module to realize global aggregation of multi-modal information.
	\item \textbf{Model IV}: \textit{Backbone $+$ IGM-Att $+$ PC-Att $+$ Sequential}. It is the sequential iterative version of our model built on Model III, where the output of the visual fusion branch serves as input to the semantic segmentation branch, enabling semantic guidance from visual fusion but without full visual-semantic collaboration.
	\item \textbf{Model V (W2F)}: \textit{Backbone $+$ IGM-Att $+$ PC-Att  $+$ Collaboration}. It is the complete version of W2F, which adds the visual-semantic collaboration on the basis of Model III, driving mutual reinforcement between visual fusion and semantic segmentation.
	\item \textbf{Model VI}: \textit{Backbone $+$ IGM-Att $+$ PC-Att from MRFS $+$ Collaboration}. This model is based on Model V, with PC-Att replaced by the PC-Att from MRFS, and all else unchanged.
	\item \textbf{Model VII}: \textit{Backbone $+$ IGM-Att $+$ PC-Att $+$ Shared Backbone}. This model is based on Model V, with the visual-semantic collaboration replaced by a shared backbone strategy between visual fusion and semantic segmentation.
\end{enumerate}

\begin{figure*}[t]
	\centering
	\includegraphics[width=1\linewidth]{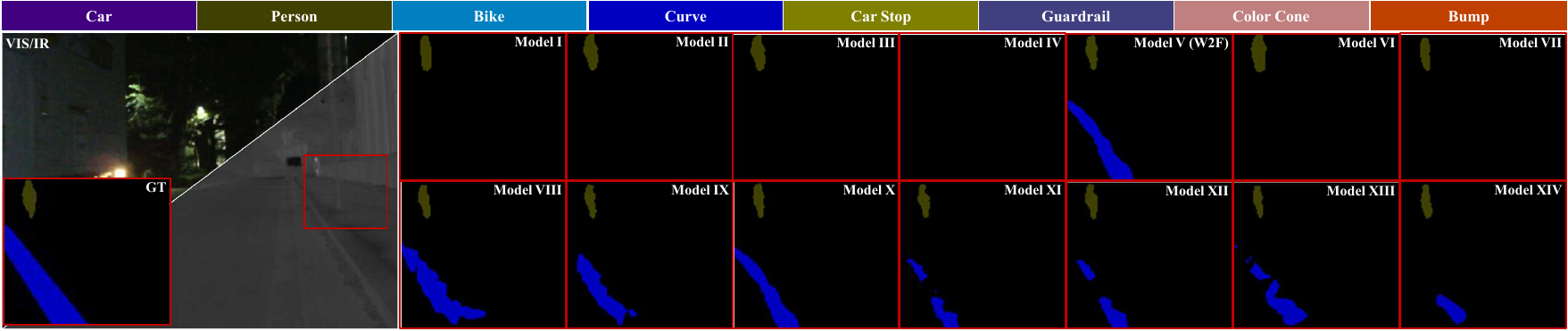}
	\vspace{-0.2in}
	\caption{Qualitative impact of each ablation variant on semantic segmentation. }\label{fig16}
\end{figure*}

\begin{table}[t]
	\caption{Quantitative segmentation of ablation studies.}\label{table-6}
	\resizebox{\linewidth}{!}{
		\setlength\tabcolsep{1.0pt}
		\footnotesize
		\centering
		\begin{tabular}{c|cccccccc|c}
			\toprule
			Model   & Car & Person & Bike & Curve & Car Stop & Guard. & Cone & Bump  & mIoU \\ \midrule
			I  &89.7    &74.3    &66.0           &48.3    &37.1   &7.8    &54.5   &51.4     &58.6             \\
			II  &90.3    &74.9    &65.6          &46.7    &44.6   &3.4    &54.3   &55.7     &59.3             \\
			III     &89.4    &75.1    &65.8           &47.6    &40.6   &12.7    &59.1   &57.1     &60.6             \\
			IV    &87.9    &72.2    &63.7          &46.9    &44.3   &15.1    &51.1   &57.9     &59.7            \\
			\rowcolor{pink!40}
			V (W2F)     &89.9    &74.7    &66.5           &47.1    &53.3   &15.8    &54.3   &53.3     &\textbf{61.5}             \\
			VI    &89.2    &74.6    &64.1           &48.2    &41.3   &10.6    &56.9  &55.3     &59.8            \\
			VII    &89.1    &74.7   &63.3           &49.1   &38.2   &6.5   &53.1  &54.7    &58.6             \\
			VIII    &89.2    &74.7    &66.3          &51.9    &39.6   &9.9    &54.4   &64.4    &61.0            \\
			IX    &89.1    &74.7    &66.2           &51.9    &39.9   &9.6    &54.6   &64.3    &61.0             \\
			X    &89.9    &74.7    &66.4           &47.1    &52.5   &16.4    &53.9   &53.5     &61.4             \\
			XI    &89.7    &74.5    &66.6           &47.2    &55.0   &9.0    &54.4   &53.3     &60.9             \\
			XII    &89.9    &74.6    &66.3           &47.1    &52.0   &14.6    &53.9   &53.5     &61.0             \\
			XIII   &89.1    &74.4    &66.2           &51.9    &38.9   &10.6    &53.6   &64.2     &60.8             \\
			XIV   &88.9    &74.7    &64.1           &48.1    &37.8   &8.2    &55.1   &55.1     &58.9             \\ \bottomrule
		\end{tabular}
	}
\end{table}

The above seven models progressively validate the gains from the ``hard'' designs: IGM-Att, PC-Att, and the collaboration mechanism. Furthermore, we conduct ablation studies on several ``soft'' designs, all based on Model V (W2F), deriving another seven model variants.
\begin{enumerate}
	\setcounter{enumi}{7}
	\item \textbf{Model VIII}: \textit{Modulating Segmentation-Guided Fusion Loss with Average Operator}. The segmentation-guided fusion loss $\mathcal{L}^{SI}_{F}$ reinforces the complementary contribution of semantic features to visual fusion, guiding visual representations to better capture semantically important regions. In this model, the maximum operator used in the designated target regions $\tau$ in the segmentation-guided fusion loss is replaced with an average operator.
	\item \textbf{Model IX}: \textit{Replacing Segmentation-Guided Fusion Loss with Global Maximum Loss}. This model replaces the segmentation-guided fusion loss with a global maximum loss, which applies the pixel-wise maximum uniformly across all regions, without differentiating the designated target regions.
	\item \textbf{Model X}: \textit{Replacing Segmentation-Guided Fusion Loss with the Proportional Maintenance Loss}. This model substitutes the segmentation-guided fusion loss with the proportional maintenance loss proposed in~\cite{zhang2020rethinking}.
	\item \textbf{Model XI}: \textit{Removing Information Restoration Prior}. The embedded information restoration prior strengthens the capability of visual fusion to extract reliable visual perceptual evidence from degraded data, thereby enhancing its ability to assist semantic segmentation. In this model, we remove the information restoration prior and use the source images as the reconstruction target for visual fusion.
	\item \textbf{Model XII}: \textit{Removing gradient correction strategy}. The gradient correction mechanism ensures better collaboration between visual fusion and semantic segmentation from an optimization perspective. For comparison, we remove the gradient correction mechanism and use loss weighting to drive multi-task joint learning.
	\item \textbf{Model XIII}: \textit{Replacing One-Zero Convolution with Standard Initialization}. The One-Zero convolution in the feature search module starts from an identity-like mapping, preserving branch-specific features while progressively learning complementary cues from the shared feature pool. In this model, we replace it with a standard randomly initialized convolution.
	\item \textbf{Model XIV}: \textit{Replacing Two-Stage Training with One-Stage Strategy}. The original W2F employs a two-stage training strategy, which first independently extracts task-specific visual and semantic representations and then refines them through collaborative optimization. This model replaces it with a single-stage strategy, jointly training all branches, feature search modules, and decoders from scratch.
\end{enumerate}
Next, we present the impact of these model variants on the performance of visual fusion and semantic segmentation tasks.

\begin{figure*}[t]
	\centering
	\includegraphics[width=1\linewidth]{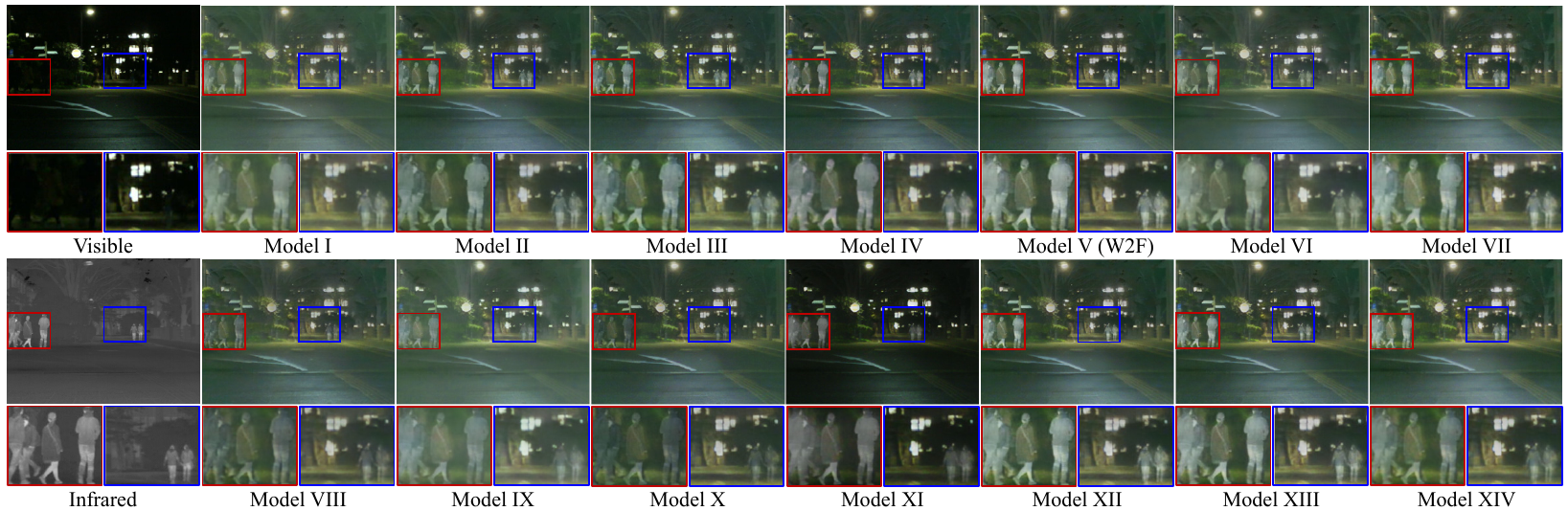}
	\vspace{-0.2in}
	\caption{Qualitative impact of each ablation variant on visual fusion.}\label{fig17-1}
\end{figure*}

\begin{figure}[t]
	\centering
	\includegraphics[width=1\linewidth]{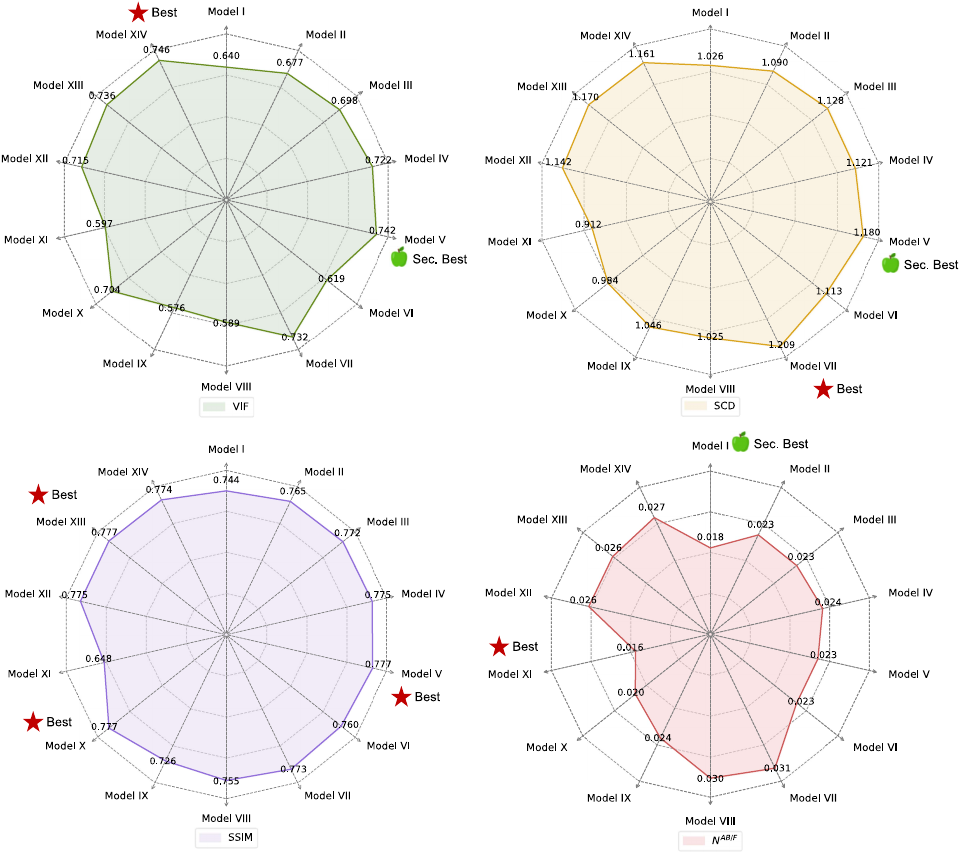}
	\vspace{-0.15in}
	\caption{Quantitative impact of each ablation variant on visual fusion. }\label{fig17-2}
\end{figure}

\subsubsection{Ablation of Semantic Segmentation}
The qualitative and quantitative ablation studies for semantic segmentation are shown in Fig.~\ref{fig16} and Table~\ref{table-6}. From Model I to Model III, the segmentation performance gradually improves, highlighting the contributions of IGM-Att and PC-Att. Model IV, which applies a sequential pipeline from visual fusion to segmentation, achieves moderate gains but remains inferior to the complete W2F model (Model V). This comparison demonstrates that our collaboration mechanism, which enables mutual reinforcement between visual fusion and segmentation, is more effective than simple sequential guidance. Replacing PC-Att with the MRFS variant in Model VI reduces segmentation performance, indicating that our PC-Att design better captures global multi-modal dependencies. Substituting the collaboration mechanism with a shared backbone in Model VII also degrades results, as sharing features without explicit cross-task alignment weakens task-specific representation learning. Models VIII$\sim$X, which modify or remove the segmentation-guided fusion loss, lead to relatively small reductions in segmentation performance, consistent with its primary effect of guiding visual fusion rather than segmentation. Removing the information restoration prior (Model XI) or the gradient correction strategy (Model XII) significantly decreases segmentation accuracy, highlighting the prior’s role in transferring visual information to segmentation and the importance of gradient correction for stable multi-task optimization. Finally, replacing the identity-initialized One-Zero convolution in the feature search module (Model XIII) or adopting a single-stage training strategy instead of two-stage training (Model XIV) reduces performance, emphasizing the value of proper initialization and staged optimization. Collectively, these results demonstrate that only the full W2F model (Model V), with all key designs, achieves superior segmentation by visual-semantic collaboration.

\begin{figure*}[t]
	\centering
	\includegraphics[width=1\linewidth]{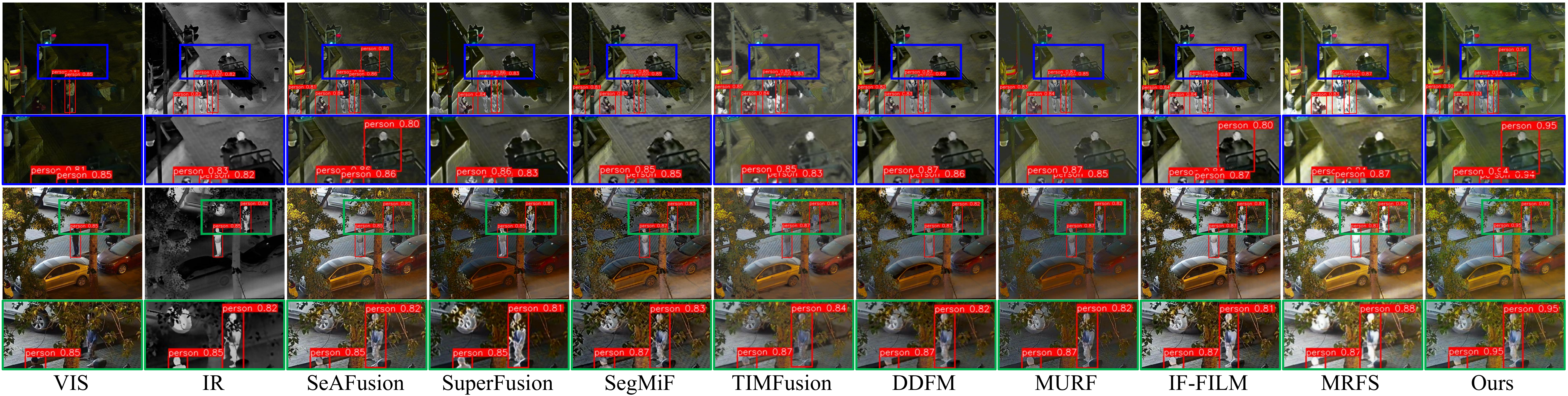}
	\vspace{-0.2in}
	\caption{Qualitative results of semantic verification with the object detection task.}\label{fig18}
\end{figure*}

\subsubsection{Ablation of Visual Fusion}
We further provide the qualitative impact of each ablation variant on visual fusion in Fig.~\ref{fig17-1}. Starting from Model I, the fused image suffers from weak foreground visibility and severe detail loss in dark regions. From Model II to Model III, the pedestrian regions gradually become clearer and the surrounding structures are better preserved, indicating that IGM-Att and PC-Att improve multi-modal information aggregation. Model IV further enhances the visibility of the pedestrians through sequential guidance, but the overall brightness and local details are still inferior to the complete W2F model (Model V). Compared with all other variants, Model V produces clearer pedestrian contours, more natural illumination, and better preservation of both foreground targets and background structures. Replacing PC-Att with the MRFS variant in Model VI slightly weakens the clarity of pedestrian regions and background lighting details. Model VII, which replaces the collaborative mechanism with a shared backbone, also leads to relatively blurred foreground regions, showing that simple feature sharing cannot fully establish effective visual-semantic interaction. Variants VIII--X, which modify the segmentation-guided fusion loss, show noticeable degradation in pedestrian saliency and local lighting details, demonstrating that the segmentation-guided fusion loss effectively guides the fused image to focus on semantically important foreground regions. After removing the information restoration prior in Model XI, the fused image becomes noticeably darker and loses details in low-light regions. Removing the gradient correction strategy in Model XII changes the appearance of the pedestrian regions and slightly affects the foreground emphasis during fusion. Model XIII, replacing the One-Zero convolution with standard initialization, slightly reduces foreground clarity, while Model XIV with one-stage training shows weaker overall contrast and detail preservation. Overall, although different variants affect visual fusion to varying degrees, the complete W2F model (Model V) consistently produces the clearest and most balanced fusion results. The quantitative comparisons in Fig.~\ref{fig17-2} further support this observation, where Model V achieves the most balanced objective performance across these metrics.

\begin{table}[t]
	\caption{Quantitative semantic verification with object detection.}\label{table-7}
	\vspace{-0.1in}
	\resizebox{\linewidth}{!}{
		\setlength\tabcolsep{1.5pt}
		\footnotesize
		\centering
		\begin{tabular}{c|cccc|c}
			\toprule
			&Precision &Recall &mAP@0.5 &mAP@0.75 &mAP@[0.5:0.95]   \\ \midrule
			VIS   &97.0       &94.1        &96.3        &73.4      &63.8                       \\
			IR &97.6       &97.3        &98.9        &88.1      &72.8                          \\
			SeAFusion    &97.7       &99.1        &99.1        &91.1      &73.4                          \\
			SuperFusion      &96.8       &98.5        & 99.1       &91.6      &73.8                        \\
			SegMiF     &98.2       &97.3        &98.8        &91.0      &73.4                         \\
			TIMFusion    &97.1       &97.6        &98.7        &90.8      &72.5                        \\  \hline
			DDFM      &97.9       &97.6        &99.0        &89.6      &73.1                        \\
			MURF      &97.4       &97.9        &98.9        &90.9      &74.2                        \\
			\rowcolor{blue!10}
			IF-FILM     &97.7       &98.8        &98.9        &91.6      &\underline{74.4}                        \\       \hline
			MRFS      & 98.2       & 97.6        & 99.3        &92.9      & 74.2                    \\
			\rowcolor{pink!40}
			Ours    & 98.5       &97.3        &98.7        &90.9      & \textbf{74.5}                        \\    	\bottomrule
		\end{tabular}
	}
	\vspace{-0.05in}
\end{table}

\subsection{Semantic Verification with Object Detection}
W2F establishes a mutual reinforcement mechanism between visual fusion and semantic segmentation, where semantic guidance enhances object visibility and distinguishability in fused images. This process can be interpreted as a form of broad semantic injection, enriching scene representations at the semantic level. Then, an interesting question is whether the injected semantics can enhance the application performance of fused images in other decision-making tasks. To investigate, we initially generate fused results using various fusion methods on the LLVIP dataset~\cite{jia2021llvip}. Subsequently, we retrain the YOLO-v5 detector using these fused images to assess their embedded semantic attributes. Each retraining involves $400$ fused images for training and $120$ for testing. Qualitative results in Fig.~\ref{fig18} show that W2F enables the most comprehensive pedestrian detection with the highest confidence. Quantitative results in Table~\ref{table-7} confirm that W2F achieves the highest mAP@[0.5:0.95], demonstrating that semantic guidance from segmentation not only improves fused image quality but also enhances their effectiveness in downstream decision-making tasks. Overall, this validates the practical value of W2F's visual-semantic collaboration beyond fusion itself.

\subsection{Complexity Discussion}
We evaluate the complexity of W2F in comparison with these visual fusion and semantic segmentation methods. Experiments are conducted on images of size $600 \times 800$, and the average running time, number of parameters, and FLOPs are summarized in Table~\ref{table-8}. As observed, W2F requires $0.15$ seconds per image, only $0.03$ seconds more than its preliminary version MRFS, while achieving substantial performance gains in both semantic segmentation and visual fusion tasks. Among methods that jointly perform visual fusion and segmentation, W2F remains competitive, being faster than SeAFusion and SegMiF. Besides, due to the dual-branch structure of W2F, a certain increase in computational overhead is unavoidable. Nevertheless, the total number of parameters and FLOPs remains moderate, indicating that the additional cost introduced by the collaboration and feature search modules is limited. Overall, considering that W2F simultaneously produces high-quality fused images and accurate semantic segmentation outputs, its complexity remains competitive among multi-modal perception methods. In the future, we will focus on improving W2F's efficiency and reducing its computational cost. One potential solution is to utilize knowledge mutual distillation, enabling visual-semantic collaboration without the need for explicit physical interaction between the task branches.

\begin{table*}[t]
	\caption{Computational analysis of multi-modal visual fusion and segmentation methods. Note: TIMFusion only reports visual fusion results because its semantic segmentation code is unavailable.}\label{table-8}
	\centering
	\resizebox{\linewidth}{!}{
		\begin{tabular}{lccccccccccccc}
			\toprule
			Method & EAEFNet & CMX & CMNeXt & CAINet & DDFM & MURF  &IF-FILM & SeAFusion & SuperFusion & SegMiF & TIMFusion & MRFS & Ours \\ \midrule
			Segmentation & \ding{52} & \ding{52} & \ding{52} & \ding{52} & \ding{56} & \ding{56} & \ding{56} & \ding{52} & \ding{52} & \ding{52} & \ding{56} & \ding{52} & \ding{52} \\
			Visual Fusion & \ding{56} & \ding{56} & \ding{56} & \ding{52} & \ding{52} & \ding{52} & \ding{52} & \ding{52} & \ding{52} & \ding{52} & \ding{52} & \ding{52} & \ding{52} \\
			Parameters (M) & 147.21 & 139.86 & 116.53 & 9.46 & 552.66 & 4.92 & 2.07 & 13.06 & 14.86 & 45.60 & 1.23 & 134.97 & 268.25 \\
			FLOPs (G) & 316.49 & 211.35 & 207.19 & 35.50 & 8700.83 & 111.78 & 140.64 & 102.53 & 142.65 & 526.20 & 337.35 & 219.16 & 394.09 \\
			Time (s) & 0.16 & 0.29 & 0.09 & 0.03 & 60.03 & 0.12 & 0.03 & 0.38 & 0.14 & 0.57 & 0.02 & 0.12 & 0.15 \\
			\bottomrule
		\end{tabular}
	}
\end{table*}

\section{Conclusion}
This paper proposes a new mutual-reinforcing paradigm for multi-modal image fusion, termed W2F, which aims to achieve simultaneous improvements in visual fusion and semantic segmentation. Firstly, W2F amplifies visual-semantic complementarity, reinforcing a solid foundation for effective collaboration. Specifically, the segmentation-guided fusion loss introduces object-aware structural guidance into visual fusion, thereby improving object visibility and distinguishability. Meanwhile, the information restoration prior strengthens the capability of visual fusion to extract reliable visual perceptual evidence from degraded observations, thereby enhancing its ability to assist the semantic branch under challenging conditions. Secondly, W2F introduces IGM-Att and PC-Att, which reinforce multi-modal information complement and aggregation from local and global perspectives, respectively, providing information support for visual-semantic collaboration. Finally, a collaborative mechanism between visual fusion and segmentation is established, bridging visual and semantic information via the feature search module, while the gradient correction strategy ensures harmonious optimization across both tasks. Extensive experiments demonstrate that W2F produces visually compelling fused images and highly accurate segmentation results, significantly enhancing perception capabilities spanning from the visual to the semantic level.

\section*{Funding}
This work was supported by the National Natural Science Foundation of China under Grant nos. 62506268 and 62276192.

\section*{Conflict of Interest}
The authors declare that they have no conflict of interest.

\section*{Ethics approval and consent to participate}
Not applicable.

\section*{Consent for publication}
Not applicable.

\section*{Data Availability}
The datasets used in this study are publicly available.

\section*{Materials availability}
Not applicable.

\section*{Code availability}
Available at \url{https://github.com/HaoZhang1018/W2F}.

\section*{Author contribution}
Hao Zhang contributed to methodology, software, writing--original draft, visualization, validation, formal analysis, investigation.
Xuhui Zuo contributed to software, visualization, validation, formal analysis.
Jiayi Ma contributed to conceptualization, methodology, supervision, funding acquisition, resources, and writing--review and editing.

\bibliographystyle{spmpsci}      
\bibliography{W2F}

\end{document}